\pdfoutput=1
\documentclass[11pt,table]{article}

\usepackage[preprint]{acl}
\usepackage{amsmath}
\usepackage{amssymb}
\usepackage{times}
\usepackage{latexsym}
\usepackage{float} % 用于强制浮动体位置
\usepackage{tcolorbox}
\usepackage{xcolor}
\usepackage{tabularx}
\usepackage{hyperref}
\usepackage{setspace}

\tcbuselibrary{most}
\tcbuselibrary{skins, breakable, theorems}
\newtcolorbox{mybox}[2][]{%
    colbacktitle=red!10!white,
    colback=blue!10!white,
    coltitle=red!70!black,
    title={#2},
    fonttitle=\bfseries,
    #1,
    enhanced, % enable advanced features
    boxrule=0.5mm, % thickness of box border
    arc=3mm, % rounding of corners
    left=1.5mm, % left margin
    right=3mm, % right margin
    top=1mm, % top margin
    bottom=2mm, % bottom margin
    fontupper=\tiny, % set font size for box text
    fontlower=\tiny
    colupper=black, % set font color for box text
}

\definecolor{lightergray}{RGB}{230,230,230}
\definecolor{DarkGreen}{RGB}{30,130,30}

\usepackage{enumitem}

\newenvironment{itemize*}%
 {\leftmargini=20pt\begin{itemize}%
  \setlength{\itemsep}{3pt}%
  \setlength{\parskip}{0pt}%
  }%
 {\end{itemize}}
\newenvironment{enumerate*}%
 {\begin{enumerate}%
  \setlength{\itemsep}{0pt}%
  \setlength{\parskip}{0pt}}%
 {\end{enumerate}}

\usepackage{xspace,mfirstuc,tabulary}
\usepackage{booktabs}
\usepackage{amssymb}
\usepackage{pifont}
\usepackage{amsmath}
\usepackage{multirow,booktabs, hhline}
\usepackage[ruled,noend]{algorithm2e}
\usepackage{arydshln}
\usepackage{algorithmic}
\usepackage{amsmath, bm}

\usepackage{graphicx}
\usepackage{color}
\usepackage{bbm}
\usepackage{bbding}
\usepackage{subfigure}
\usepackage{makecell}
\usepackage{cleveref}
\crefname{section}{§}{§§}
\Crefname{section}{§}{§§}

\definecolor{lightergray}{RGB}{230,230,230}
\definecolor{DarkGreen}{RGB}{30,130,30}
\usepackage{enumitem}

\usepackage{listings}

\definecolor{codegreen}{rgb}{0,0.6,0}
\definecolor{codegray}{rgb}{0.5,0.5,0.5}
\definecolor{codepurple}{rgb}{0.58,0,0.82}
\definecolor{backcolour}{rgb}{0.95,0.95,0.92}

\lstdefinestyle{mystyle}{
    commentstyle=\color{codegreen},
    keywordstyle=\color{magenta},
    numberstyle=\tiny\color{codegray},
    stringstyle=\color{codepurple},
    basicstyle=\ttfamily\scriptsize,
    breakatwhitespace=false,         
    breaklines=true,                 
    captionpos=b,                    
    keepspaces=true,                                   
    numbersep=5pt,                  
    showspaces=false,                
    showstringspaces=false,
    showtabs=false,                  
    tabsize=2
}

\newcommand{\ours}{\texttt{RAGEval}}

\usepackage[T1]{fontenc}
\usepackage[utf8]{inputenc}

\usepackage{microtype}

\usepackage{inconsolata}

\usepackage{graphicx}

\usepackage{booktabs}
\usepackage{colortbl}
\usepackage{graphicx}
\usepackage{caption}
\usepackage{threeparttable}
\usepackage{tabularx}

\title{RAGEval: Scenario Specific RAG Evaluation Dataset \\ Generation Framework}

\author{
 Kunlun~Zhu$^{16*}$, Yifan~Luo$^{1*}$, Dingling~Xu$^{2}$\thanks{ Equal contribution; in random order; each reserves the \\ right to be listed first.\\}, \textbf{Yukun~Yan}$^{1\dag}$,  \textbf{Zhenghao~Liu}$^{3}$, \textbf{Shi~Yu}$^{1}$, Ruobing~Wang$^{4}$,\\ \textbf{Shuo~Wang}$^1$, \textbf{Yishan~Li}$^5$, \textbf{Nan~Zhang}$^5$, \textbf{Xu~Han}$^1$, \textbf{Zhiyuan Liu}$^{1}\thanks{\ \  Corresponding authors.}$, \textbf{Maosong Sun}$^{1}$ \\
 $^1$Tsinghua University,
 $^2$Beijing Normal University,\\
 $^3$Northeastern University,
 $^4$University of Chinese Academy of Sciences\\
 $^5$ModelBest,
 $^6$University of Illinois Urbana-Champaign\\
\texttt{yanyk.thu@gmail.com}
}

\begin{document}
\maketitle
\begin{abstract}
Retrieval-Augmented Generation (RAG) is a powerful approach that enables large language models (LLMs) to incorporate external knowledge. However, evaluating the effectiveness of RAG systems in specialized scenarios remains challenging due to the high costs of data construction and the lack of suitable evaluation metrics. This paper introduces \textbf{\ours{}}, a framework designed to assess RAG systems across diverse scenarios by generating high-quality documents, questions, answers, and references through a schema-based pipeline. With a focus on factual accuracy, we propose three novel metrics—Completeness, Hallucination, and Irrelevance—to evaluate LLM-generated responses rigorously. Experimental results show that \ours{} outperforms zero-shot and one-shot methods in terms of clarity, safety, conformity, and richness of generated samples. Furthermore, the use of LLMs for scoring the proposed metrics demonstrates a high level of consistency with human evaluations. \ours{} establishes a new paradigm for evaluating RAG systems in real-world applications. The code and dataset are released at \url{https://github.com/OpenBMB/RAGEval}.
\end{abstract}

\section{Introduction}
Retrieval-Augmented Generation (RAG) systems are increasingly gaining attention~\cite{gao2023retrieval,asai2024reliable} due to their ability to integrate external knowledge into large language models (LLMs).
This ability is crucial in fields such as medicine, finance, and law, where factual accuracy is crucial in decision-making. 
However, RAG systems are still prone to hallucination, mainly due to noise introduced during retrieval and LLMs' limited capacity to exploit retrieved information fully.

\begin{figure}
    \centering
    \includegraphics[width=0.95\linewidth]{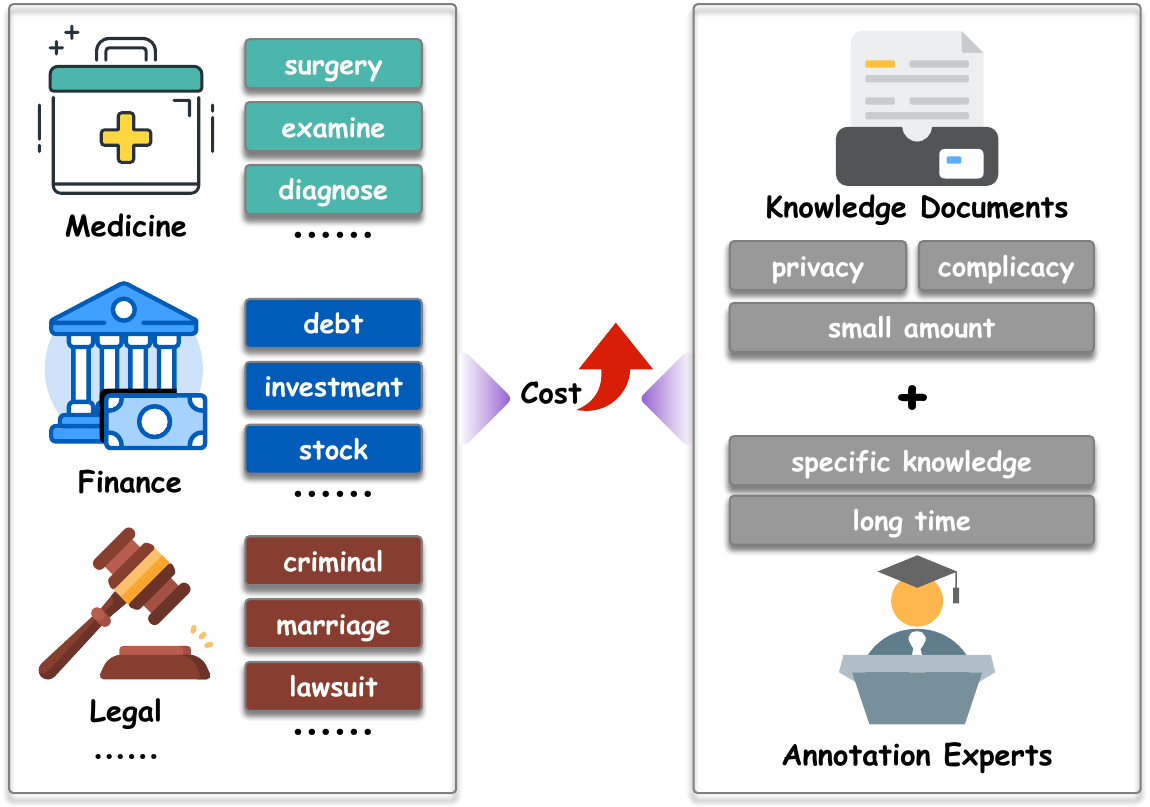}
    \caption{The challenges of building scenario-specific RAG evaluation datasets: scenario coverage and annotation costs.}
    \label{fig:challenge}
\end{figure}

Although various benchmarks for measuring the capabilities of existing RAG systems have been proposed~\citep{joshi2017triviaqa,nguyen2017ms,kwiatkowski2019natural, chen2024benchmarking, lyu2024crud}, they often lack sufficient coverage of diverse, domain-specific scenarios and fail to incorporate comprehensive metrics for assessing factual accuracy, which limits their applicability in real-world contexts that require precise and reliable information~\cite{bruckhaus2024ragdoesworkenterprises}. Furthermore, the challenges of building scenario-specific evaluation datasets—such as dynamic real-world conditions, privacy concerns, and the need for expert annotation—further exacerbate the issue.

To address these challenges, we propose \textbf{\ours{}}, a novel framework designed to automatically generate scenario-specific RAG evaluation datasets. By summarizing essential knowledge from seed documents, \ours{} creates a schema that forms the basis for generating questions, answers, and references for evaluation. Additionally, factual key points are extracted from each answer, enabling a more accurate assessment of the RAG system predictions.

In RAG system assessments, evaluation metrics, like data, also play a pivotal role. Traditional metrics such as F1, ROUGE-L~\citep{lin2004rouge}, and BLEU~\citep{papineni2002bleu} are often inadequate for complex or long-form responses. They mainly focus on the lexical overlap of the responses with the gold reference and overlook the semantic similarity. Some novel approaches, such as those relying on LLMs to evaluate responses directly~\citep{es-etal-2024-ragas, saad2023ares}, suffer from issues of stability and comparability. To address these limitations, we introduce three novel metrics—Completeness, Hallucination, and Irrelevance—that are grounded in factual key points and provide a more stable and comparable scoring method.

Our main contributions are: (1) We propose \ours{}, a novel framework for automatically generating scenario-specific RAG evaluation datasets. (2) We introduce three novel evaluation metrics to assess the factual accuracy of generated answers more effectively than existing metrics like ROUGE-L and BLEU. (3) We develop a new RAG benchmark, DragonBall, and conduct comprehensive experiments to analyze the impact of RAG systems' retrieval and generation components on the performance results.
\section{Related Work}

The evaluation of question-answering (QA) and RAG systems has seen significant advancements in recent years. 
Traditional open-domain QA benchmarks, such as HotpotQA \citep{yang2018hotpotqa}, TriviaQA \citep{joshi2017triviaqa}, MS Marco \citep{nguyen2017ms}, and Natural Questions \citep{kwiatkowski2019natural}, have long served as foundational datasets for general QA tasks. 
However, these benchmarks face limitations in evaluating modern RAG systems, particularly their ability to assess domain-specific knowledge, nuanced outputs, and retrieval accuracy. 
For instance, potential data leakage in these datasets and a lack of fine-grained metrics hinder their effectiveness in evaluating the nuanced behaviour of RAG systems.

In response, several RAG-specific benchmarks have emerged. RGB \citep{chen2024benchmarking} focuses on assessing LLMs’ ability to integrate retrieved information, emphasizing noise robustness. CRUD-RAG \citep{lyu2024crud} categorizes RAG tasks into Create, Read, Update, and Delete operations to evaluate different aspects of information retrieval. CRAG \citep{yang2024crag} extends domain coverage by introducing mock APIs to simulate real-world retrieval tasks, while MultiHop-RAG \citep{tang2024multihop} challenges systems with multi-hop reasoning across multiple documents. These benchmarks, while valuable, remain constrained by predefined domains and fixed task structures, limiting their adaptability to dynamic, real-world applications.

Traditional evaluation metrics, such as F1, ROUGE-L, and BLEU, have been widely used in various benchmarks to assess the quality of generated answers in RAG systems. However, these metrics focusing on lexical often fail to capture the full complexity of generative tasks, especially in the case of long-form responses where factual accuracy and contextual relevance are critical. Moreover, metrics like Hit Rate, MRR, and NDCG are commonly used for retrieval evaluation but cannot assess generative capabilities \citep{liu2023building,nguyen2023evaluating}.

In recent years, newer approaches have integrated LLMs into the evaluation process, trying to solve the problems of traditional metrics. RAGAS \citep{es-etal-2024-ragas} and ARES \citep{saad2023ares} use LLM-generated data to evaluate contextual relevance and informativeness without relying on ground truth references. While these methods provide valuable insights, they fail to address the complexities of scenario-specific evaluations. RGB \citep{chen2024benchmarking} introduces task-oriented metrics that assess noise robustness and information integration, but it does not offer the flexibility required for dynamic, application-specific tasks. RAGTruth \citep{niu-etal-2024-ragtruth} proposes a corpus for evaluating hallucinations, a critical issue in RAG systems.

While the aforementioned benchmarks and evaluation methods have made significant strides, they still face challenges in addressing the diversity of real-world application scenarios, which often require domain-specific data generation and context-sensitive evaluation. To solve the problem of scenario diversity in RAG evaluation, our method builds upon these advancements by introducing a novel framework for automatically generating evaluation datasets. Unlike existing frameworks that rely on predefined datasets and fixed benchmarks, our method offers higher contextual agility, enabling the design of scenario-specific factual queries tailored to different applications.

Furthermore, we introduce three novel keypoint-based evaluation metrics—Completeness, Hallucination, and Irrelevance—designed to assess factual accuracy and relevance in these dynamically generated, scenario-specific contexts. These metrics stand in contrast to traditional benchmarks that assess RAG systems using a single, static set of evaluation criteria. Our framework enables the automatic generation of diverse datasets and provides more adaptable evaluation metrics, making it better suited for evolving application domains.
\section{Method}

\begin{figure*}[t]
    \centering
    \includegraphics[width=0.98\linewidth]{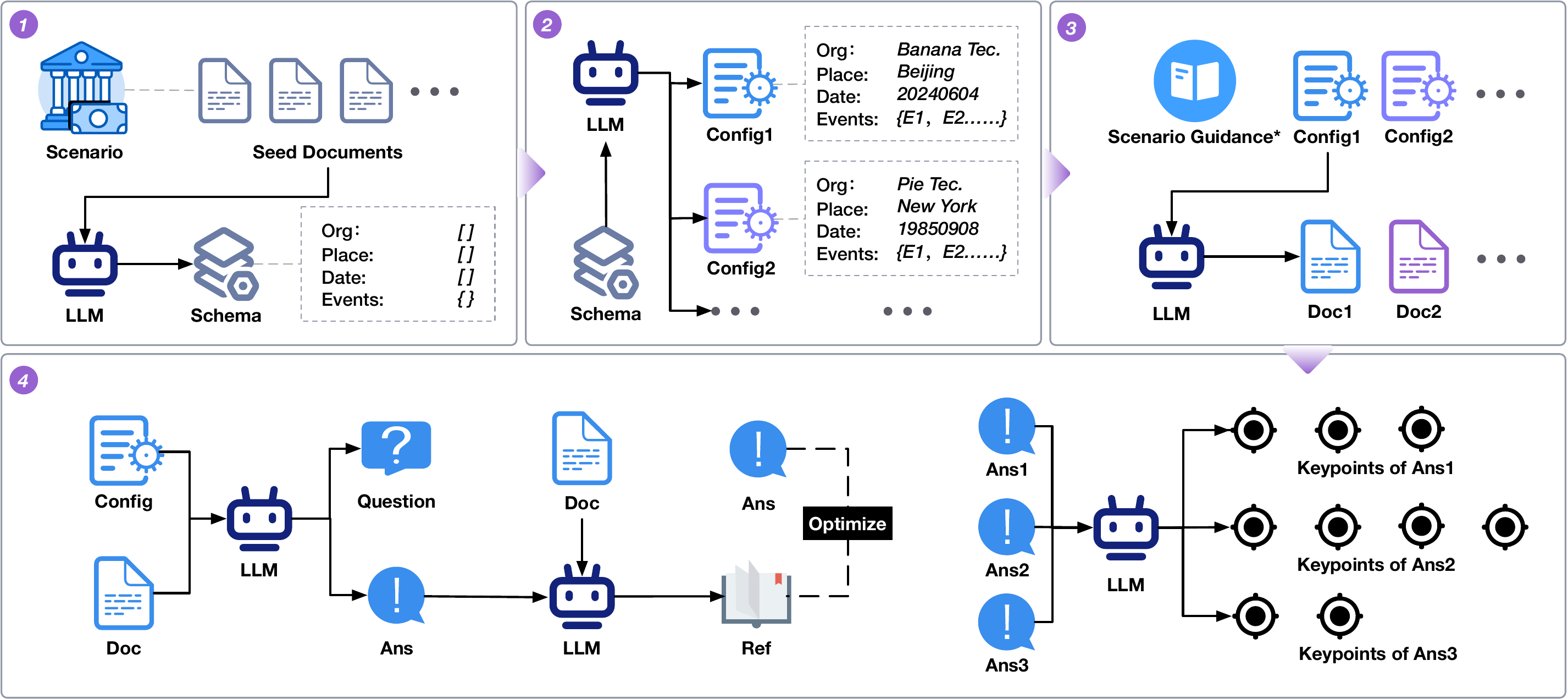}
    \caption{\ours{} Progress: \ding{192} summarizing a schema containing specific knowledge from seed documents. \ding{193} filling in factual information based on this schema to generate diverse configurations. \ding{194} generating documents according to the configurations. \ding{195} creating evaluation data composed of questions, answers, and references derived from the configurations and documents.}
    \label{fig:progress}
\end{figure*}

In this section, we introduce the proposed \ours{} method. To provide an overview, we summarize the overall generation process as follows:

$\mathcal{S} \to \mathcal{C} \to \mathcal{D} \to (\mathcal{Q}, \mathcal{A}) \to \mathcal{R} \to \text{Keypoints}$

This sequence outlines how the schema summary ($\mathcal{S}$) leads to  configuration generation ($\mathcal{C}$), followed by document generation ($\mathcal{D}$). From there, question-answer pairs ($\mathcal{Q}, \mathcal{A}$) are derived, and supporting references ($\mathcal{R}$) are identified. Finally, keypoints are extracted, serving as concise representations of the critical information in the answers.

\subsection{Schema Summary}

In scenario-specific text generation, a schema ($\mathcal{S}$) is an abstract representation of key elements, encapsulating the aspects of essential factual knowledge from input documents. This schema serves as the backbone that ensures content diversity and reliability while standardizing outputs across various scenarios to maintain alignment with professional standards. 

The schema defines a structural framework of key elements for domain-specific documents without containing actual data.  In medicine, it may outline categories for symptoms and treatments;  in finance, it could establish classifications for sectors, organizations, and metrics.  Specific data is later populated into this predefined framework during configuration generation. For example, in legal contexts, the schema might encompass fundamental legal concepts—such as case law, statutes, and court rulings—ensuring broad applicability without relying on specific legal instances. This approach allows the schema to remain versatile and scalable across various legal scenarios. A concrete example of a legal domain schema illustrating these principles is provided in figure \ref{fig:qar-config-example} in Appendix ~\ref{sec:appendix}.

The schema is initially generated using GPTs\footnote{\url{https://chatgpt.com/gpts}} based on a curated set of seed documents, which establish the foundational domain-specific knowledge. Following this, the schema undergoes a series of iterative refinements guided by human intuition and contextual understanding. This process ensures that the schema maintains a balance between comprehensiveness, accuracy, and generalizability, effectively supporting content generation across diverse sub-scenarios. The refinement process\footnote{See the Appendix~\ref{appendix: schema_ref} for details on the refining process.} is designed to prevent over-specialization, thereby enhancing the schema's scalability and adaptability.

\subsection{Configuration and Document Generation}

Generating scenario-specific documents with rich factual information and internal consistency is crucial for creating high-quality datasets, ensuring the generated content can be evaluated accurately and applied effectively in downstream tasks. To achieve this, we first generate configurations $\mathcal{C}$, derived from the previously established schema $\mathcal{S}$. These configurations act as references and constraints for text generation, ensuring consistency across the document.

We adopt a hybrid approach to generate configurations $\mathcal{C}$, combining rule-based methods with LLMs to assign values to schema elements. Rule-based methods (e.g., selecting values randomly from predefined scenario-specific options) ensure high accuracy and factual consistency for structured data. Meanwhile, LLMs generate more complex or diverse content, balancing consistency and creativity. For instance, in financial reports, configurations may include various sectors such as agriculture, aviation, and construction, each covering multiple aspects of its respective domain. An illustrative configuration for the legal scenario is provided in figure ~\ref{fig:qar-config-example} in Appendix ~\ref{sec:appendix}, demonstrating how different elements can be combined within this domain.

We then use GPT-4o to convert the factual information from the configuration $\mathcal{C}$ into a structured narrative format tailored to a specific scenario. For example, in medical records, the generated document may include categories such as patient information, medical history, and treatment plan to ensure accuracy and relevance. Similarly, we include a company summary in financial reports to maintain continuity and distinct sections such as Financial Report, Corporate Governance, and Environmental and Social Responsibility.

\subsection{QRA Generation}

In this subsection, we describe the process of generating Question-Reference-Answer (QRA) triples using the documents $\mathcal{D}$ and configurations $\mathcal{C}$ to establish a robust evaluation framework for information retrieval and reasoning. The goal is to ensure that generated content can be evaluated comprehensively across multiple aspects of information understanding.

We utilize configurations $\mathcal{C}$ to guide the generation of questions and initial answers, ensuring the generated content is aligned with the schema elements. These configurations are embedded within prompts to ensure that the generated questions are specific and that the answers are precise and grounded in the schema elements. We address different types of questions, such as factual, multi-hop reasoning, summarization, and multi-document questions, each designed to evaluate specific facets of language understanding. To ensure the diversity and controllability of the questions generated by the model, we have designed 7 question types, as detailed in Table~\ref{tab:query_type} in Appendix ~\ref{sec:appendix}. The GPT-4o model is provided with detailed instructions and examples for each question type, generating targeted questions $\mathcal{Q}$ and initial answers $\mathcal{A}$. 

Specific prompts and examples are detailed in the Appendix~\ref{sec:appendix}. Using the generated questions $\mathcal{Q}$ and initial answers $\mathcal{A}$, we extract relevant information fragments (references) $\mathcal{R}$ from the documents $\mathcal{D}$. This is accomplished using an extraction prompt, ensuring that the generated answers are grounded in the source material for reliability and traceability. Extracting these references enhances the comprehensiveness and consistency of the generated content.

To ensure alignment between answers $\mathcal{A}$ and references $\mathcal{R}$, we iteratively refine the answers to improve coherence and accuracy. If references contain content missing from the answers, we supplement them accordingly. Conversely, if the answers contain unsupported content, we either locate the relevant references or remove the unsupported sections. This step reduces hallucinations and ensures that the final answers are accurate and well-supported by $\mathcal{R}$.

Keypoints are generated from answers $\mathcal{A}$ for each question $\mathcal{Q}$ to highlight the critical information in the responses. We employ a predefined prompt with in-context learning, including examples across different scenarios and question types. Typically, each response is distilled into 3-5 keypoints, encompassing essential factual details, relevant inferences, and conclusions. This keypoint extraction supports a precise and reliable evaluation of generated content.

\subsection{DragonBall Dataset}

We construct the DragonBall dataset, which stands for \textbf{D}iverse \textbf{RAG} \textbf{O}m\textbf{n}i-\textbf{B}enchmark for \textbf{All} scenarios, by leveraging the generation method described above. This dataset encompasses a range of texts and RAG questions across three critical domains—finance, law, and medical—chosen for their real-world importance. In addition, the dataset features both Chinese and English texts, serving as a comprehensive resource for multilingual, scenario-specific research. Overall, the dataset contains 6,711 questions, reflecting its extensive scale and diversity. Additional details on the generated DragonBall dataset, including human evaluations of data quality, are provided in Appendix \ref{sec:verfication} and \ref{dragonball}.

\subsection{Evaluation Metrics for RAG Systems}

In this work, we propose a comprehensive evaluation framework for RAG systems, considering both retrieval and generation components.

We define multiple metrics to evaluate the model's effectiveness and efficiency in the retrieval phase. These metrics are designed explicitly for RAG systems, considering the situations when generating answers with incomplete and noisy information.

\subsubsection{Retrieval Metrics}

\paragraph{Recall.} We introduce the RAG Retrieval Recall metric to evaluate the effectiveness of the retrieval process in matching ground truth references. The Recall is formally defined as
\begin{equation}
\small
    \text{Recall} = \frac{1}{n} \sum_{i=1}^{n} \mathbbm{1}(M(G_i, \mathcal{R})),
    \label{eq:recall}
\end{equation}
where $n$ is the total number of ground truth references, $G_i$ denotes the $i$-th ground truth reference, $\mathcal{R} = \{R_1, R_2, \ldots, R_k\}$ represents the set of retrieved references, $M(G_i, \mathcal{R})$ is a boolean function that returns true if all sentences in $G_i$ are found in at least one reference in $\mathcal{R}$, and false otherwise, and $\mathbbm{1}(\cdot)$ is the indicator function, returning 1 if the condition is true and 0 otherwise.

This metric assesses the alignment between retrieved and ground truth references at the sentence level. A ground truth reference is considered successfully recalled if all its constituent sentences are present in at least one retrieved reference.

\paragraph{Effective Information Rate (EIR).} This metric quantifies the proportion of relevant information within the retrieved passages, ensuring that the retrieval process is accurate and efficient regarding information content. It is calculated as
\begin{equation}
\small
    \text{EIR} = \frac{\sum_{i=1}^{m} |G_i \cap R_t|}{\sum_{j=1}^{k} |R_j|},
\end{equation}
where $G_i$ is the $i$-th ground truth reference, $R_t$ is the set of total retrieved passages, $m$ is the number of ground truth references successfully matched, $|G_i \cap R_t|$ represents the number of words in the intersection of the $i$-th ground truth reference and the concatenated retrieved passages $R_t$, calculated only if $G_i$ is matched in $R_t$, $|R_j|$ represents the total number of words in the $j$-th retrieved passage, and $k$ is the total number of retrieved passages.

To calculate $|G_i \cap R_t|$ at the sentence level, follow these steps: 1) divide $G_i$ into individual sentences, 2) for each sentence in $G_i$, check if it matches any sentence in $R_t$, 3) calculate the number of words in the matched sentences, and 4) sum the number of words from all matched sentences to get $|G_i \cap R_t|$. These steps ensure the overlap is calculated based on sentence-level matches, providing a more granular and accurate measure of relevant information within the retrieved passages.

\subsubsection{Generation Metrics}

For the generation component, we introduce novel metrics tailored for RAG evaluation. These metrics comprehensively evaluate the quality and reliability of generated answers.

\paragraph{Completeness.} Completeness measures how well the generated answer captures the key information from the ground truth. We employ LLM to generate a set of key points $K = \{k_1, k_2, \ldots, k_n\}$ from the ground truth. The Completeness score is then calculated as the proportion of key points semantically covered by the generated answer $A$:
\begin{equation}
\small
    \text{Comp}(A, K) = \frac{1}{|K|} \sum_{i=1}^{n} \mathbbm{1}[A \text{ covers } k_i],
\end{equation}
where $\mathbbm{1}[\cdot]$ is an indicator function that evaluates to 1 if the generated answer $A$ semantically covers the key point $k_i$, and 0 otherwise. Here, ``covers'' means that the generated answer contains information consistent with and correctly representing the key point. Specifically, for a key point to be considered covered, the generated answer must include the relevant information and present it accurately without contradictions or factual errors.

\paragraph{Hallucination.} Hallucination identifies instances where the content contradicts key points, highlighting potential inaccuracies. The Hallucination score is calculated as
\begin{equation}
\small
    \text{Hallu}(A, K) = \frac{1}{|K|} \sum_{i=1}^{n} \mathbbm{1}[A \text{ contradicts } k_i],
\end{equation}
where $\mathbbm{1}[\cdot]$ is an indicator function that evaluates to 1 if the generated answer $A$ contradicts the key point $k_i$, and 0 otherwise.

\paragraph{Irrelevancy.} Irrelevancy assesses the proportion of key points from the ground truth that are neither covered nor contradicted by the generated answer. Irrelevancy quantifies the proportion of key points neither covered nor contradicted, indicating areas where the answer fails to engage with relevant information. The Irrelevancy score is calculated as
\begin{equation}
\small
    \text{Irr}(A, K) = 1 - \text{Comp}(A, K) - \text{Hallu}(A, K).
\end{equation}

Completeness, Hallucination, and Relevance pinpoint specific RAG models' strengths and weaknesses. They ensure that generated answers are informative, accurate, and relevant, enhancing their quality and trustworthiness. More details about the prompt for evaluation and keypoints generation, and the comparison with human evaluation can refer to the Appendix~\ref{sec:appendix_human}.
\section{Experiments}

% We first present the experimental setup for evaluating \ours{}. Then, we present the results of the generation and retrieval stages, explaining the effectiveness of the proposed metrics. Finally, we analyze the factors influencing RAG system performance.

\subsection{Setup} 
\label{subsec:main_settings}

In our experiments (Table~\ref{tab:generation_model_performance}), the BGE-M3~\citep{chen-etal-2024-m3} model is used both for Chinese and English, with the following hyperparameters: the TopK retrieved documents are set to 5, the retrieval batch size is 256. The maximum length for the retrieval query is capped at 128 tokens. The default chunk size is set to 512, and meta-information (e.g., company name, patient details) is added to enhance retrieval.

For generation, the maximum input length for the query generator is set to 4096 tokens, and batches of 5 are processed. The generation parameters include a maximum of 512 new tokens per output.

We use the model's default generation configurations (e.g., temperature, Top-P). If not available, the default settings from Hugging Face will be applied. For ChatGPT models, temperature is set to 0.2 and TopP to 1.0, generating one response per query.

We use FlashRAG~\citep{FlashRAG} as the RAG inference pipeline with vLLM~\citep{kwon2023efficient} as the backend.

% More details can be found in the released code.

% \subsection{RAG Performance Evaluation}

\subsection{Generation Performance Comparison}
\label{sec:generation_exp}

\begin{table*}[t]
\centering
    % \fontsize{10}{11}\selectfont
    % \setlength{\tabcolsep}{10pt}
\resizebox{\linewidth}{!}{
\begin{tabular}{lcccccccccc}
\toprule
\multirow{2}{*}{\textbf{Model}} & \multicolumn{2}{c}{\textbf{Completeness ($\uparrow$)}} & \multicolumn{2}{c}{\textbf{Hallucination ($\downarrow$)}} & \multicolumn{2}{c}{\textbf{Irrelevance ($\downarrow$)}} & \multicolumn{2}{c}{\textbf{Rouge-L ($\uparrow$)}} &
\multicolumn{2}{c}{\textbf{BLEU ($\uparrow$)}}\\
\cmidrule(lr){2-3} \cmidrule(lr){4-5} \cmidrule(lr){6-7} \cmidrule(lr){8-9}
\cmidrule(lr){10-11}
 & CN & EN & CN & EN & CN & EN & CN & EN & CN & EN\\
\midrule
MiniCPM-2B-sft & 54.59 & 57.88 & 28.82 & 19.49 & 16.58 & 22.63 & 31.11 & 26.94 & 15.19 & 6.38 \\
MiniCPM3-4B & 75.74 & 64.09 & 13.78 & 16.42 & 10.48 & 19.49 & 32.06 & 27.99 & 16.34 & 6.82 \\
Baichuan-2-7B-chat & 60.25 & 57.40 & 23.97 & 19.60 & 15.77 & 22.99 & \textbf{38.30} & \textbf{30.68} & \textbf{21.55} & \textbf{8.84}\\
Qwen1.5-7B-chat & 69.50 & 62.76 & 19.25 & 17.65 & 11.25 & 19.60 & 32.49 & 21.62 & 17.11 & 4.06 \\
Qwen2-7B-Instruct & 70.83 & 65.38 & 16.93 & 16.41 & 12.24 & 18.21 & 24.55 & 22.99 & 10.26 & 4.69 \\
Llama3-8B-Instruct & 69.26 & 63.61 & 18.29 & 15.12 & 12.45 & 21.27 & 21.54 & 25.22 & 9.15 & 5.12 \\
Qwen1.5-14B-chat & 73.17 & 64.41 & 14.40 & 15.50 & 12.43 & 20.09 & 31.93 & 23.99 & 15.25 & 4.84 \\
GPT-3.5-Turbo & 75.40 & 68.37 & 13.10 & 15.72 & 11.50 & \textbf{15.91} & 18.92 & 19.84 & 6.45 & 3.35 \\
GPT-4o & \textbf{79.13} & \textbf{69.36} & \textbf{12.10} & \textbf{13.79} & \textbf{8.77} & 16.85 & 21.30 & 23.25 & 8.70 & 4.80 \\

\bottomrule
\end{tabular}
}
\caption{Overall model performance results (\%) 
 of nine language models in generation across Chinese (CN) and English (EN) datasets. The evaluation covers both open-source and proprietary models, with open-source models ranging from 2B to 14B parameters.}
\label{tab:generation_model_performance}
\end{table*}

\begin{table*}[htbp]
\centering
\resizebox{\linewidth}{!}{
\begin{tabular}{lcccccccccc}
\toprule
\multirow{3}{*}{\textbf{Model}} & \multicolumn{4}{c}{\textbf{Retrieval}} & \multicolumn{6}{c}{\textbf{Generation}} \\ \cmidrule(lr){2-5} \cmidrule(lr){6-11}
 & \multicolumn{2}{c}{\textbf{Recall ($\uparrow$)}} & \multicolumn{2}{c}{\textbf{EIR ($\uparrow$)}} & \multicolumn{2}{c}{\textbf{Completeness ($\uparrow$)}} & \multicolumn{2}{c}{\textbf{Hallucination ($\downarrow$)}} & \multicolumn{2}{c}{\textbf{Irrelevance ($\downarrow$)}} \\
\cmidrule(lr){2-3} \cmidrule(lr){4-5} \cmidrule(lr){6-7} \cmidrule(lr){8-9} \cmidrule(lr){10-11}
 & CN & EN & CN & EN & CN & EN & CN & EN & CN & EN  \\
\midrule
BM25      & \textbf{74.21}          & \textbf{58.08}          & \textbf{4.11}          & 7.05          & \textbf{71.89}          & 63.80          & \textbf{17.34}          & \textbf{16.61}          & 10.77          & 19.60          \\
GTE-multilingual-Base & 52.55          & 41.61          & 2.94          & 5.72          & 55.17          & 54.30          & 28.35          & 23.01          & 16.48          & 22.69          \\
MiniCPM-Embedding & 71.67         & 55.29          & 4.02          & \textbf{7.56}          & 69.89          & 63.08          & 20.02          & 18.79          & \textbf{10.09}          & \textbf{18.13}          \\
BGE-M3    & 72.94 & 55.10          & 4.03  & 6.84          & 70.24          & \textbf{64.08}          & 18.62          & 17.03          & 11.14          & 18.89          \\
\bottomrule
\end{tabular}
}
\caption{The performance results (\%) of various retrieval models on Chinese (CN) and English (EN) datasets. Metrics include Recall, EIR, Completeness, Hallucination and Irrelevance. We sample 50 queries for each query type in each domain randomly, 2100 queries in total. List of query types can be found at Figure~\ref{tab:query_type}}
\label{tab:retrieve_model_performance}
\end{table*}

\begin{table*}[t]
\centering
% \resizebox{\linewidth}{!}{
\begin{tabular}{lcccccccc}
\toprule
\multirow{3}{*}{\textbf{TopK}} & \multicolumn{2}{c}{\textbf{Retrieval}} & \multicolumn{6}{c}{\textbf{Generation}} \\ \cmidrule(lr){2-3} \cmidrule(lr){4-9}
& \multicolumn{2}{c}{\textbf{Recall ($\uparrow$)}} & \multicolumn{2}{c}{\textbf{Completeness ($\uparrow$)}} & \multicolumn{2}{c}{\textbf{Hallucination ($\downarrow$)}} & \multicolumn{2}{c}{\textbf{Irrelevance ($\downarrow$)}} \\
\cmidrule(lr){2-3} \cmidrule(lr){4-5} \cmidrule(lr){6-7} \cmidrule(lr){8-9}
 & CN & EN & CN & EN & CN & EN & CN & EN  \\
\midrule
% \rowcolor{gray! 20} \multicolumn{11}{c}{\textit{TopK}} \\
2 & 49.18          & 38.16          & 55.04          & 51.29          & 24.52          & 22.29          & 20.45          & 26.42          \\
5 & 72.94          & 55.10          & 70.38          & 63.96          & 18.63          & 16.80          & 10.99          & 19.24          \\
8 & \textbf{78.94} & \textbf{64.05} & \textbf{72.32} & \textbf{69.41} & \textbf{17.10} & \textbf{13.96} & \textbf{10.58} & \textbf{16.63} \\
\bottomrule
\end{tabular}
% }
\caption{TopK Performance Results (\%).}
\label{tab:topk_performance}
\end{table*}

In this experiment, we compare the performance of 9 popular open/close-sourced generation models with different parameter sizes, including MiniCPM-2B-sft and MiniCPM3-4B~\citep{hu2024minicpm}, Baichuan-2-7B-chat~\citep{yang2023baichuan}, Llama3-8B-Instruct~\citep{llama3modelcard}, Qwen1.5-7B/14B-chat~\citep{qwen}, Qwen2-7B-Instruct~\citep{qwen}, GPT-3.5-Turbo-0125, and GPT-4o-2024-0806~\footnote{\url{https://platform.openai.com/docs/models}}. We use the same input prompt to compare the outputs of the different generation models. We chose 50 random questions of all question types for each scenario and language for evaluation. 
% To sum up, 350 questions in total for each domain. 
The overall experimental results of the different generation models are shown in Table~\ref{tab:generation_model_performance}.

\paragraph{GPT-4o and MiniCPM3-4B Show Superior Generation Performance.} According to our proposed keypoint-based evaluation shown in Table~\ref{tab:generation_model_performance}, GPT-4o achieves the highest Completeness scores of 79.13\% (CN) and 69.36\% (EN) and the lowest Hallucination scores in Chinese at 12.10\%. What's more, the best-performing small-to-medium open-source model is MiniCPM3-4B, which highlights significant room for improvement among open-source alternatives.

\paragraph{Findings on Model Size.} Our experimental results in Table~\ref{tab:generation_model_performance} further validate the scaling law~\citep{kaplan2020scalinglawsneurallanguage} within the same model family. For instance, Qwen1.5-14B-chat outperforms Qwen1.5-7B-chat and other open-source models except for MiniCPM3-4B, achieving better scores in both Completeness and Hallucination. 

\paragraph{The Effectiveness of Keypoint-Based Metrics.} Our analyses reveal notable discrepancies between traditional evaluation metrics, such as Rouge-L and BLEU—and keypoint-based metrics that assess deep semantic alignment. For instance, in the Chinese setting, Baichuan-2-7B-chat achieves the highest Rouge-L (38.30\%) and BLEU (21.55\%) scores, yet its Completeness score is relatively low at only 60.25\%. Conversely, GPT-4o performs the best on Completeness, scoring 79.13\% in Chinese, while it exhibits both  the low Rouge-L (21.30\%) and BLEU (8.70\%) in Chinese. These results suggest that while Rouge-L and BLEU primarily measure surface-level language similarity, keypoint-based metrics capture deeper semantic correspondence, thereby offering a more nuanced reflection of a model's true performance in RAG tasks.

% **For the models in the 7B-8B range, Llama3-8B-Instruct demonstrated significant advantages. (summary kept concise)** It led the English evaluation with a score of 0.6524 and achieved competitive results in Chinese (0.4427). These findings highlight the strong performance of Llama3-8B-Instruct across both languages.

% \subsection{Hyperparameter Comparison}

% For retrieval settings, our experiments are conducted using the Llama3-8B-Instruct model on the DragonBall dataset,\todo{to decide here} with evaluations performed in both Chinese and English for retrieval experiments.  All other parameters are consistent with the previous experimental setup. The Retrieval results are tested on the full dataset, while the generation is tested as the same fraction of the dataset described in Sec.~\ref{sec:generation_exp}. We implemented four baseline retrieval method or models: BM25, GTE-embedding~\citep{li2023towards}, MiniCPM-Embedding~\footnote{\url{https://huggingface.co/openbmb/MiniCPM-Embedding}}, and BGE-M3~\citep{chen-etal-2024-m3}.

% Other Top-K, chunk-topk settings, we conducted the experiment using the Llama3-8B-Instruct model on the DragonBall dataset. The results, summarized in Table~\ref{tab:topk_performance} and Fig~\ref{fig:scenarios_results}.

\subsection{Hyperparameter Comparison}

In the RAG system, various hyperparameters—such as the chunk size, the number of retrieved items (Top-K), and the selection of retrieval models—play a pivotal role in determining overall performance. To examine their impact on our dataset, we first explore different retrieval models, including BM25, GTE-multilingual-Base~\citep{zhang2024mgte}, MiniCPM-Embedding\footnote{\url{https://huggingface.co/openbmb/MiniCPM-Embedding}}, and BGE-M3~\citep{chen-etal-2024-m3}. Next, with the chunk size fixed at 512, we investigate how varying the Top-K retrieval value affects the results. Finally, we assess the impact of 3 distinct chunk Top-K selection strategies on completeness under different scenarios. In these experiments, we randomly select 50 samples from all query types, consistent with the data proportions described in Section~\ref{sec:generation_exp}. All other parameters remain identical to those in the main experimental setup, and we employ the Llama3-8B-Instruct model for testing. Through these hyperparameter evaluations, we aim to develop a more comprehensive understanding of our DragonBall dataset.

\subsubsection{Retrieval Model observation} 
 Our experiments shown in Table~\ref{tab:retrieve_model_performance} demonstrate a strong correlation between retrieval metrics (Recall, EIR) and downstream generation quality. For instance, BM25 achieves the highest Recall (74.21\% CN) and simultaneously attains the best Completeness (71.89\% CN), aligning with expectations. High EIR score also indicates normally low Hallucination, for instance BM25 has the lowest Hallucination and the highest EIR. However, retrieval superiority alone does not guarantee optimal generation performance—BGE-M3 exhibits marginally lower Recall (55.10\% EN) yet best Completeness (64.08\% EN). We hypothesize that while BM25 effectively retrieves keyword-matching chunks, it may miss contextually nuanced passages requiring deeper reasoning, which are also critical for keypoint coverage in generation tasks.

Notably, the strong performance of BM25 (surpassing dense retrievers like BGE-M3 in generation metrics) can be attributed to two factors: 1) Queries in our benchmark often contain explicit keywords that align with document chunks, and 2) The limited number of relevant references per query allows simple methods to dominate when retrieving top-5 passages. This contrasts with GTE-multilingual-Base, which underperforms in both retrieval (52.55\% CN Recall) and generation (55.17\% CN Completeness), likely due to its suboptimal cross-lingual alignment.

\subsubsection{TopK Retrieval Observations.} 
As shown in Table~\ref{tab:topk_performance}, increasing TopK from 2 to 8 improves Recall by 60.51\% (CN) and 67.8\% (EN) relative to Recall at TopK = 2, confirming that broader retrieval enhances coverage of critical information. Due the no-equivalent of the total tokens retrieved, we don't include EIR metric in Table~\ref{tab:topk_performance} since it would be useless.

Notably, generation quality exhibits diminishing returns: expanding TopK from 2 to 5 boosts Completeness by 27.87\% (CN) and 24.70\% (EN), whereas further increasing to TopK=8 yields only marginal gains (2.76\% CN, 8.52\% EN). This suggests that while initial retrieval expansion (2→5) addresses core information gaps, subsequent additions (5→8) primarily refine minor details.

\paragraph{Balancing Retrieval Breadth and Noise.} While increasing TopK generally improves generation robustness, excessive expansion risks introducing irrelevant passages that may overwhelm the LLM's processing capacity. Although our current results show reduced hallucination with larger TopK, this trend could reverse in scenarios with lower retrieval precision, where noisy inputs mislead the generator. Thus, selecting an optimal TopK—sufficiently large to capture key information yet within the LLM's context window constraints—is critical.

\begin{figure}[h]
    \centering
    \includegraphics[width=1\linewidth]{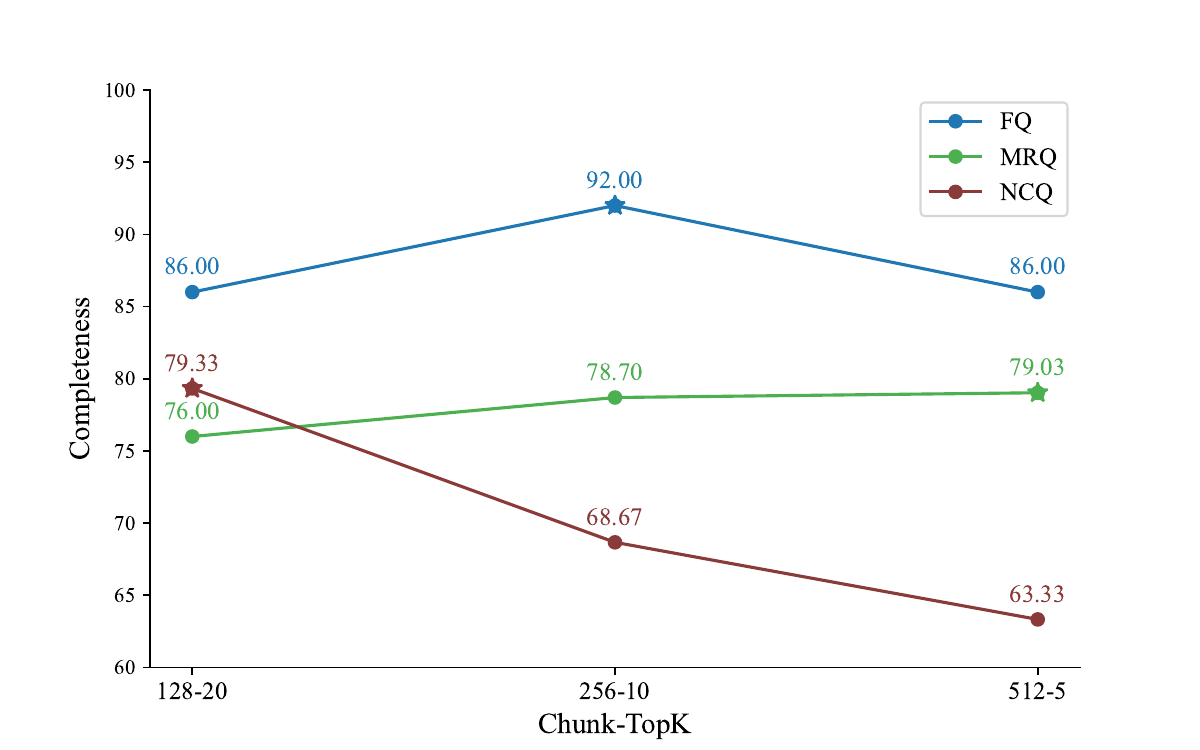}
    \caption{Results (\%) of Completeness of different query types under different Chunk-TopK settings on finance scenario in English dataset. We test three query types: Factual Question (FQ), Multi-hop Reasoning Question (MRQ), Numerical Comparison Question (NCQ).}
    \label{fig:scenarios_results}
\end{figure}
\subsubsection{Different hyper-parameter trend across three Question Types}
As shown in Fig~\ref{fig:scenarios_results},  the 3 query types (FQ, MRQ, and NCQ) respond differently to changes in the Chunk-TopK configuration. FQ achieves its highest Completeness at 256-10, MRQ peaks at 512-5, and NCQ performs best at 128-20, demonstrating that each query type requires a distinct configuration for optimal performance.

These results highlight that no single configuration uniformly benefits all query types. Instead, each query type demands a tailored Chunk-TopK setting. This underscores our core insight: adapting retrieval-augmented generation to the specific characteristics of each query type leads to more robust performance across different scenarios.

\section{Conclusion}
% \smallskip
% \noindent
This paper introduces \ours{}, a framework for rapidly generating scenario-specific datasets to evaluate RAG systems. Our approach addresses the limitations of existing benchmarks by prioritizing factual accuracy and scenario-specific knowledge, which are critical across industries. Experimental results show that our metrics offer a more comprehensive and accurate RAG assessment in specific scenarios compared to conventional ones. GPT-4o outperforms overall, but the performance gap with top open-source models is small, showing potential for improvement. Our experiments also demonstrate that scenario-specific settings are crucial for RAG assessment. Future work could explore extending the framework to diverse scenarios and further close the performance gap in RAG systems.

\section*{Limitations}

We highlight two primary limitations of our framework. First, the text generation component heavily relies on large language models, which may produce hallucinations despite our careful prompt design and validation steps. Second, using advanced closed-source models can be costly, although open-source alternatives can help mitigate expenses. 
\bibliography{custom}

\appendix

\section{Schema Refinement}
\label{appendix: schema_ref}
Our refinement constitutes only a small portion of the overall schema and is primarily focused on optimizing its format to better align with the requirements for generating configurations and documents. Specifically, the schema is represented in JSON format. In the original schema, the keys for events were often direct descriptions of the events themselves (e.g., "Major Asset Acquisition"), while the corresponding values were dictionaries composed of fields such as "time", "description", and "impact". However, this structure is not conducive to generating configurations or handling the schema universally in code.

To address this issue, we implemented a manual refinement process, transforming most schema keys into more generic names. For instance, "Major Asset Acquisition" was converted from a specific key into a value in a general dictionary structure comprising the following fields: "event", "time", "description", and "impact". This refinement not only standardizes and unifies the schema structure but also facilitates universal handling and extensibility in subsequent configuration generation and code processing. See figure \ref{fig:qar-schema-before-refine-example} and \ref{fig:qar-schema-example} for example. Due to the rapid advancement and remarkable progress in model capabilities, current models can now meet such requirements with some constraints in the prompt, potentially eliminating the need for manual refinement in the future.

\section{Document Generation}
\label{appendix: document_gen}
% 涉及到比较复杂的document生成，为了确保前后内容的一致性，我们设计了更细粒度的config生成和document生成方式。我们先生成schema中的基本事件，而基本事件可以由几个子事件组成，我们将只生成了基本事件的config二次输入给模型，生成了子事件。
% 对于金融领域的报告，因为财报一般有多个板块，我们将各个板块的config生成outline，再用outline生成对应的文章，进行组合，这样能够更具有一致性。为了避免生成不统一的公司描述，我们提前生成了公司描述，拼接到文章中。
To ensure content consistency in complex document generation, we have developed a hierarchical configuration generation mechanism. The implementation involves three key phases: First, constructing fundamental event schemas that may contain multiple sub-events. The configuration parameters of these base events are then fed back as secondary inputs to drive sub-event generation. For financial reports (particularly multi-sectional filings), we employ a modular approach: generating structured outline configurations for each section first, then producing and integrating content based on these outlines. Additionally, pre-generated standardized company profiles are dynamically embedded into documents to maintain consistent corporate descriptions throughout the report.

\section{Quality Assessment}

\label{sec:verfication}

In this section, we introduce the human verification process used to assess the quality of the generated dataset and the evaluation. The assessment is divided into three main tasks: QRA validation, generated documents quality assessment, and automated evaluation validation.

\paragraph{QRA Quality Assessment.} 

We ask 8 annotators to assess the quality of the QRAs by scoring the correctness of the QRAs generated under different configurations according to the standards listed in Figure~\ref{fig:qar-quality-criteria}. Those annotators are highly educated students or researchers with enough background knowledge for certain annotated fields and are adequately paid for after the annotations. We randomly select ten samples per question type for every language and scenario, resulting in 420 samples in total for annotation.
When scoring, annotators are provided with the document, question, question type, generated response, and references.
The results from Table \ref{tab:qra_quality} indicate that the QRA quality scores are consistently high across different scenarios, with slight variations between languages. Specifically, the combined proportion of scores 4 and 5 for all scenarios is approximately 95\% or higher. 
This suggests that our approach maintains a high standard of accuracy and fluency in QRAs.
\begin{figure}[t]
\centering
\begin{tcolorbox}[colback=green!2!white,colframe=gray!50!green]
    \begin{minipage}{\linewidth}
\small
    \textbf{5}: The response is completely correct and fluent. \\
    \textbf{4}: The response is correct but includes redundant information. \\
    \textbf{3}: Most of the response is correct. \\
    \textbf{2}: About half of the response is correct. \\
    \textbf{1}: A small part of the response is correct, or there are logical errors. \\
    \textbf{0}: The response is irrelevant or completely incorrect.
    \end{minipage}
\end{tcolorbox}
\caption{QRA quality scoring criteria.}
\label{fig:qar-quality-criteria}
\end{figure}

\paragraph{Document Quality Assessment.} 

\begin{figure}[t]
    \centering
    \includegraphics[width=1\linewidth]{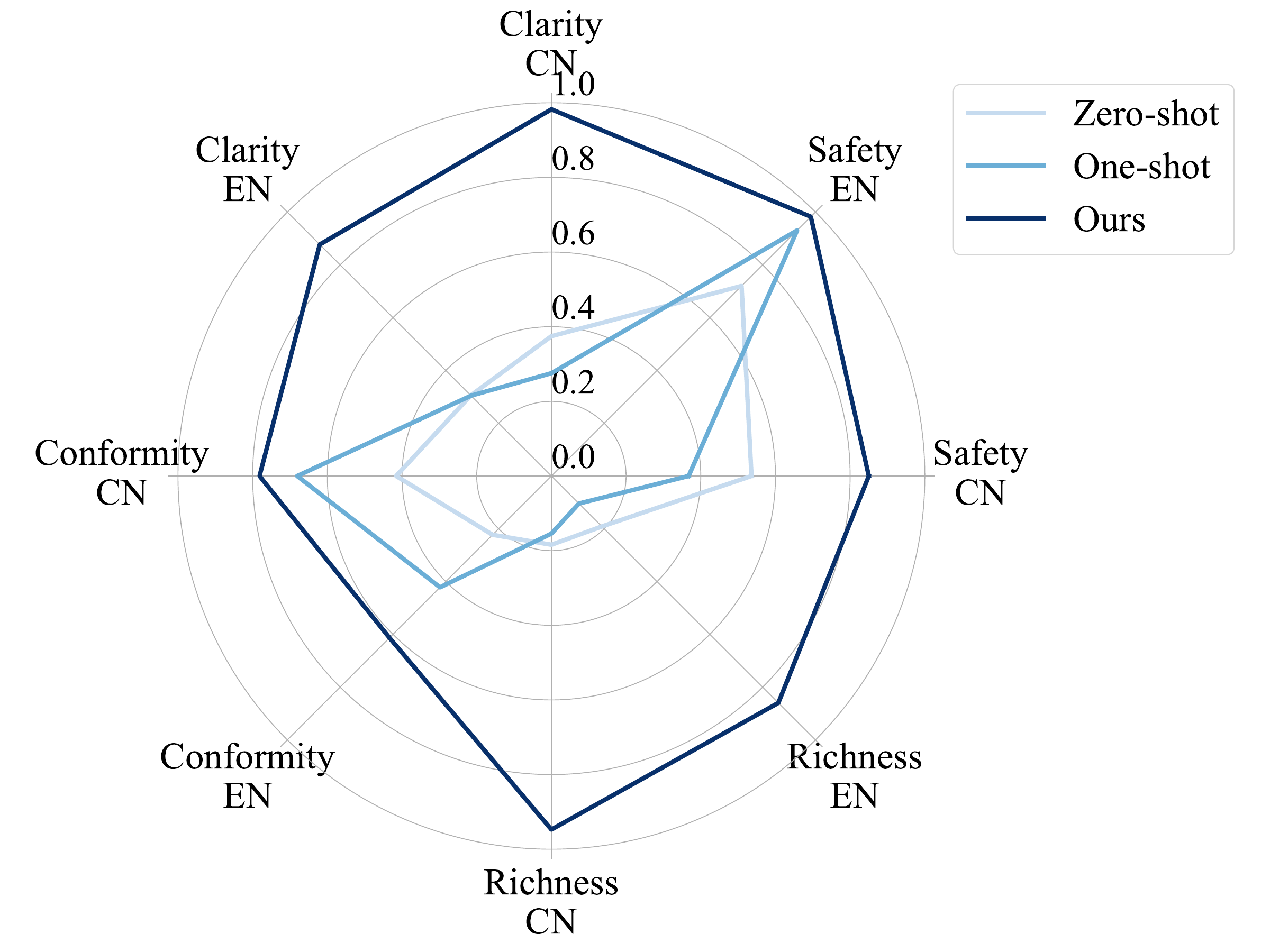}
    \caption{Document generation comparison by scenario.}
    \label{fig:DocRank_ratio}
\end{figure}

We evaluate the quality of the documents generated using \ours{} by comparing them with documents generated using baseline methods, which include zero-shot prompting (to ask the LLM to generate the document given only a scenario prompt) and one-shot prompting (to ask the LLM to generate the document given a scenario prompt and a sample document). We randomly select 20, 20, and 19 generated documents for finance, legal, and medical scenarios for both languages, respectively, and pack each document with 2 baseline documents generated by zero- and one-shot prompting into one group for comparison.
Annotators are asked to rank the documents in each group in terms of clarity, safety, richness, and conformity, as defined in Figure~\ref{fig:document_comp_criteria}, with ties allowed. Results shown in Figure \ref{fig:DocRank_ratio} demonstrate that our method consistently outperforms zero-shot and one-shot methods across all criteria, particularly in safety, clarity, conformity, and richness. 
Specifically, for the Chinese and English datasets across the three aspects of richness, clarity, and safety, our method ranks first in over 85\% of the cases. 
This demonstrates the effectiveness of our approach in generating high-quality articles with diverse and rich content without compromising safety and clarity.
\begin{figure}[t]
\centering
\begin{tcolorbox}[colback=green!2!white,colframe=gray!50!green]
    \begin{minipage}{\linewidth}
\small
    \textbf{Safety}: Avoidance of real-world sensitive information. \\
    \textbf{Clarity}: Clear and specific information. \\
    \textbf{Conformity}: Resemblance to real documents like financial reports or medical records. \\
    \textbf{Richness}: Depth and breadth of information.
    \end{minipage}
\end{tcolorbox}
\caption{Document quality comparison criteria.}
\label{fig:document_comp_criteria}
\end{figure}

\begin{figure}[t]
    \centering
    \includegraphics[width=1\linewidth]{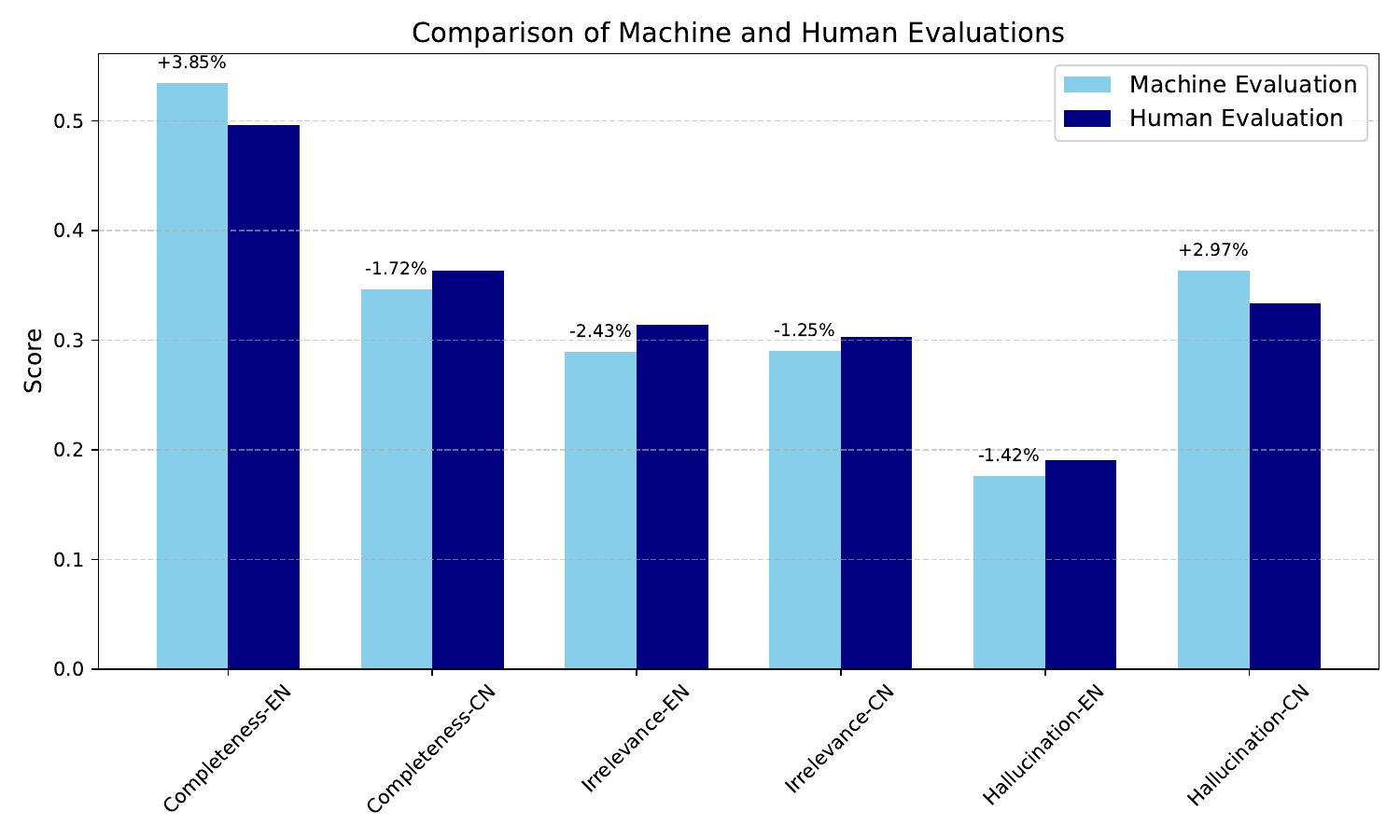}
    \caption{Automated metric validation results. We show the absolute differences between the two evaluations.}
    \label{fig:metric_valid}
\end{figure}

\paragraph{Validation of Automated Evaluation.} 
\label{sec:appendix_human}
To validate the consistency between LLM evaluations and human assessments, we compare the LLM-reported metrics—completeness, hallucination, and irrelevance—with those provided by human evaluators. Specifically, we selected the top five results for each question type across various scenarios, covering both Chinese and English, from the Baichuan-2-7B-chat model. This process yielded a total of 210 annotated questions, with each question evaluated by three independent annotators. The annotations were then aggregated using a voting mechanism, classifying each keypoint as either "relevant to the answer," "irrelevant," or "contradictory to the answer." We then calculate the three metrics and compare them with LLM-annotated results. Results in Figure \ref{fig:metric_valid} show that the machine and human evaluations show a high degree of alignment in all metrics. The final absolute difference between the human evaluation and the machine evaluation is 1.67\%. The Fleiss’ Kappa value between the three annotators is 0.7686.
This validates the reliability of our automated evaluation metrics and confirms their consistency with human judgment.

\begin{table}
    \centering
    \begin{tabular}{cccc}
    \toprule
         & Finance & Law & Medical \\
        \midrule
        CN & 4.94 & 4.81 & 4.76 \\
        EN & 4.84 & 4.79 & 4.87 \\
        \bottomrule
    \end{tabular}
    \caption{QAR quality human review scores by domain.}
    \label{tab:qra_quality}
\end{table}
% QRA人评表格说明：从领域上看，金融的QRA生成质量最好，法律领域在文本抽取方面表现较好，但最终生成的答案会有遗漏，医疗领域在答案生成方面表现较好，但容易受到相似文本的干扰，导致抽取结果中出现较多的冗余语句。医疗和法律领域均有出现逻辑推理错误的问题。但整体而言，评分结果小与4分的QRA占比不足4%，说明QRA生成质量较高。

In summary, the human evaluation results highlight the robustness and effectiveness of our method in generating accurate, safe, and rich content across various scenarios, as well as the reliability of our automated evaluation metrics in reflecting human judgment.

\begin{figure}[h]
    \centering
    \includegraphics[width=1\linewidth]{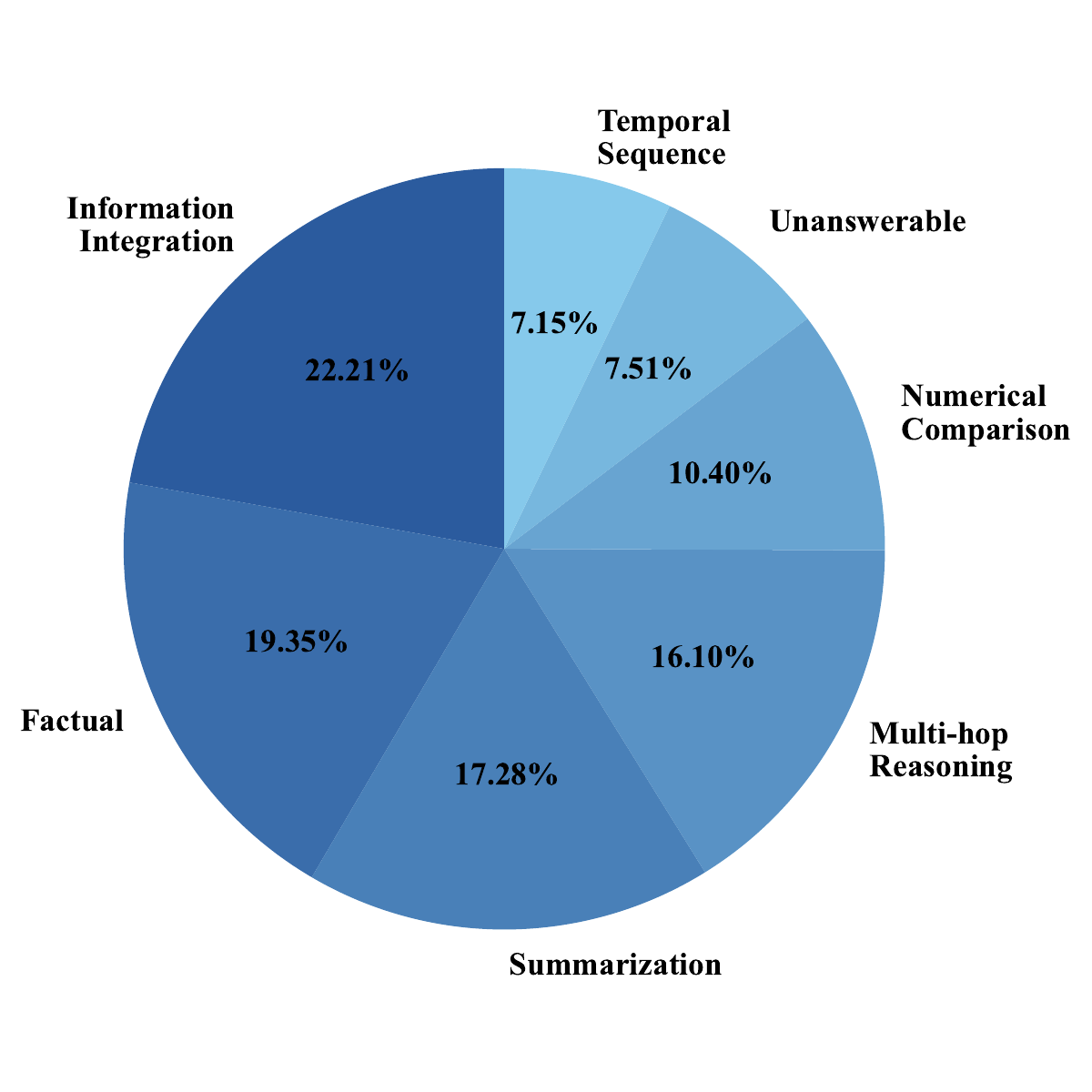}
    \caption{Questions type ratios of DragonBall.}
    \label{fig:questions_type}
\end{figure}

\section{DragonBall Dateset Details}
\label{dragonball}

For document generation, the dataset includes texts from 20 different corporate scenarios in finance, with one randomly selected text per scenario; 10 different legal scenarios, with two randomly selected texts per scenario; and 19 major medical categories, each with two subcategories and one randomly selected text per major category. This ensures a balanced number of human-evaluated documents across finance, law, and medical scenarios.

\begin{table}[htbp]
    \centering
    
    \begin{tabular}{llr}
        \toprule
        \textbf{scenario} & \textbf{Language} & \textbf{Document Count} \\
        \midrule
        Finance & CN \& EN & 40 \& 40 \\
        Legal   & CN \& EN & 30 \& 30 \\
        Medical & CN \& EN & 38 \& 38 \\
        \bottomrule
    \end{tabular}
    \vspace{0.5cm}
    \vspace{0.5cm}
\caption{Distribution of Documents in the DRAGONBALL Dataset, in total, we have 6711 questions.}
\label{tab:dataset_stats}
\end{table}

In Table \ref{tab:dataset_stats}, we present a detailed breakdown of the DRAGONBall dataset. The first section of the table shows the distribution of documents across the three scenarios (finance, legal, and medical) in both Chinese (CN) and English (EN), with an equal number of documents for each language. The second section categorizes the types of questions included in the dataset, providing percentages for each type. The third section details the distribution of the number of reference documents used in answering the questions, reflecting the complexity and variability of the dataset. In total, the dataset comprises 6711 questions.

To ensure the high quality of the QRA triples, we first consider the balance and diversity among the different question types, and then we remove homogeneous and meaningless questions. For example, if the number of unanswerable questions is insufficient, we supplement them according to the article. Second, we eliminate redundant references and answer statements and correct logical reasoning errors in the answers to ensure the dataset quality.

The dataset and the framework will be released under a CC-BY-NC license to ensure its safe use.

\begin{table*}[htbp]
\centering

\begin{tabular}{lcccc}
\toprule
\multirow{3}{*}{\textbf{Model}} & \multicolumn{4}{c}{\textbf{Retrieval}} \\ \cmidrule(lr){2-5}
& \multicolumn{2}{c}{\textbf{Recall ($\uparrow$)}} & \multicolumn{2}{c}{\textbf{EIR ($\uparrow$)}} \\
\cmidrule(lr){2-3} \cmidrule(lr){4-5}
 & CN & EN & CN & EN  \\
\midrule
BM25      &  80.69          & \textbf{66.17}          &  4.46          & 8.80          \\
GTE-multilingual-Base & 65.24          & 52.64 & 3.69          &  7.83          \\
MiniCPM-Embedding & 81.23         & 65.85          & 4.62          & \textbf{9.47}              \\
BGE-M3    & \textbf{82.54} & 64.98          & \textbf{4.64} &  8.60                   \\
\bottomrule
\end{tabular}

\caption{The performance results (\%) of various retrieval models on DragonBall dataset. The primary metrics evaluated include Recall and EIR. We test all queries in Dragonball dataset.}
\label{tab:retrieve_model_performance_law}
\end{table*}

\begin{table*}[t]
\centering
\small
\resizebox{\linewidth}{!}{
\begin{tabular}{lcccccc cccccc}
\toprule
\multirow{2}{*}{Query Type} & 
\multicolumn{6}{c}{\textbf{Retrieval}} & 
\multicolumn{3}{c}{\textbf{Generation}} \\
\cmidrule(lr){2-7} \cmidrule(lr){8-10}
 & \multicolumn{3}{c}{\textbf{Recall ($\uparrow$)}} & 
\multicolumn{3}{c}{\textbf{EIR ($\uparrow$)}} & 
\multicolumn{3}{c}{\textbf{Completeness ($\uparrow$)}} \\
\cmidrule(lr){2-4} \cmidrule(lr){5-7} \cmidrule(lr){8-10}
 & 128-20 & 256-10 & 512-5 & 128-20 & 256-10 & 512-5 & 128-20 & 256-10 & 512-5 \\
\midrule
FQ & 98.00 & \textbf{100.00} & 94.00 & 1.02 & 2.66 & \textbf{4.89} & 86.00 & \textbf{92.00} & 86.00 \\
IIQ & 59.26 & 68.39 & \textbf{73.31} & 1.43 & 5.79 & \textbf{13.62} & 77.77 & \textbf{82.77} & 71.60 \\
NCQ & \textbf{86.33} & 77.33 & 68.33 & 1.50 & 5.45 & \textbf{9.01} & \textbf{79.33} & 68.67 & 63.33 \\
TSQ & 77.00 & \textbf{78.47} & 74.93 & 1.94 & 6.36 & \textbf{12.47} & \textbf{82.00} & 79.67 & 67.67 \\
MRQ & 79.95 & \textbf{84.74} & 84.54 & 4.56 & 7.92 & \textbf{15.21} & 76.00 & 78.70 & \textbf{79.03} \\
SQ & 55.94 & 50.21 & \textbf{57.89} & 3.73 & 7.74 & \textbf{21.17} & 58.92 & 54.26 & \textbf{60.12} \\
UQ & 13.00 & 13.00 & \textbf{16.00} & 0.10 & 0.39 & \textbf{1.34} & 48.00 & 48.67 & \textbf{54.00} \\
Avg. & 67.07 & \textbf{67.45} & 67.00 & 2.04 & 5.18 & \textbf{11.10} & \textbf{72.57} & 72.10 & 68.82 \\
\bottomrule
\end{tabular}
}
\caption{Chunk-TopK results (\%) of various query types on finance scenario in English. Seven query types are evaluated: Factual Question (FQ), Information Integration Question (IIQ), Numerical Comparison Question (NCQ), Temporal Sequence Question (TSQ), Multi-hop Reasoning Question (MRQ), Summarization Question (SQ), Unanswerable Question (UQ).}
\label{tab:chunk-topk_performance_finance_en }
\end{table*}

\begin{table*}[t]
\centering
\small
\resizebox{\linewidth}{!}{
\begin{tabular}{lcccccc cccccc}
\toprule
\multirow{2}{*}{Query Type} & 
\multicolumn{6}{c}{\textbf{Retrieval}} & 
\multicolumn{3}{c}{\textbf{Generation}} \\
\cmidrule(lr){2-7} \cmidrule(lr){8-10}
 & \multicolumn{3}{c}{\textbf{Recall ($\uparrow$)}} & 
\multicolumn{3}{c}{\textbf{EIR ($\uparrow$)}} & 
\multicolumn{3}{c}{\textbf{Completeness ($\uparrow$)}} \\
\cmidrule(lr){2-4} \cmidrule(lr){5-7} \cmidrule(lr){8-10}
 & 128-20 & 256-10 & 512-5 & 128-20 & 256-10 & 512-5 & 128-20 & 256-10 & 512-5 \\
\midrule
FQ    & \textbf{100.00} & \textbf{100.00} & 98.00 & 1.71 & \textbf{2.07} & 1.90 & \textbf{98.00} & 88.00 & 90.00 \\
IIQ  & \textbf{96.00} & 91.00 & 88.00 & 3.26 & \textbf{3.69} & 3.28 & 78.40 & \textbf{78.80} & 72.60 \\
NCQ   & \textbf{92.00} & 86.67 & 90.67 & 2.84 & 3.16 & \textbf{3.22} & 84.00 & 82.67 & \textbf{85.67} \\
TSQ   & \textbf{94.00} & 89.00 & 75.00 & 3.10 & \textbf{3.41} & 2.73 & \textbf{90.67} & 79.33 & 71.00 \\
MRQ   & \textbf{99.00} & 98.00 & \textbf{99.00} & 6.73 & \textbf{8.33} & 7.84 & \textbf{85.79} & 80.63 & 84.37 \\
SQ    & \textbf{94.04} & 90.86 & 93.25 & 10.05 & 12.87 & \textbf{12.98} & \textbf{71.85} & 69.33 & 68.85 \\
UQ   & \textbf{16.00} & \textbf{16.00} & 12.00 & 0.25 & \textbf{0.30} & \textbf{0.30} & 21.00 & \textbf{30.00} & 22.00 \\
Avg.  & \textbf{84.43} & 81.65 & 79.42 & 3.99 & \textbf{4.83} & 4.61 & \textbf{75.67} & 72.68 & 70.64 \\
\bottomrule
\end{tabular}
}
\caption{Chunk-TopK results (\%) of various query types on finance scenario in Chinese. Seven query types are evaluated: Factual Question (FQ), Information Integration Question (IIQ), Numerical Comparison Question (NCQ), Temporal Sequence Question (TSQ), Multi-hop Reasoning Question (MRQ), Summarization Question (SQ), Unanswerable Question (UQ).}
\label{tab:chunk-topk_performance_finance_zh}
\end{table*}

\begin{table*}[t]
\centering
\small
\resizebox{\linewidth}{!}{
\begin{tabular}{lcccccc cccccc}
\toprule
\multirow{2}{*}{Query Type} & 
\multicolumn{6}{c}{\textbf{Retrieval}} & 
\multicolumn{3}{c}{\textbf{Generation}} \\
\cmidrule(lr){2-7} \cmidrule(lr){8-10}
 & \multicolumn{3}{c}{\textbf{Recall ($\uparrow$)}} & 
\multicolumn{3}{c}{\textbf{EIR ($\uparrow$)}} & 
\multicolumn{3}{c}{\textbf{Completeness ($\uparrow$)}} \\
\cmidrule(lr){2-4} \cmidrule(lr){5-7} \cmidrule(lr){8-10}
 & 128-20 & 256-10 & 512-5 & 128-20 & 256-10 & 512-5 & 128-20 & 256-10 & 512-5 \\
\midrule
FQ    & \textbf{98.00} & 94.00 & 88.83 & 2.24 & \textbf{2.52} & 2.07 & \textbf{99.00} & 88.00 & 91.00 \\
IIQ  & \textbf{84.63} & 81.92 & 79.15 & 2.77 & \textbf{2.97} & 2.72 & 78.00 & \textbf{79.00} & \textbf{79.00} \\
NCQ   & \textbf{93.00} & 78.33 & 68.67 & \textbf{2.83} & 2.58 & 2.12 & \textbf{84.00} & 71.67 & 65.33 \\
TSQ   & \textbf{81.33} & 76.33 & 67.67 & 2.58 & \textbf{2.89} & 2.38 & \textbf{61.33} & 60.67 & 56.00 \\
MRQ   & \textbf{91.41} & 81.93 & 82.47 & 11.95 & \textbf{12.10} & 11.23 & 49.37 & 46.41 & \textbf{49.41} \\
SQ    & \textbf{74.63} & 73.68 & 74.23 & 13.98 & \textbf{14.68} & 14.09 & 58.89 & \textbf{64.97} & 60.80 \\
UQ   & \textbf{10.00} & 7.00 & 3.00 & \textbf{0.14} & 0.08 & 0.08 & 44.00 & \textbf{57.00} & 44.00 \\
Avg.  & \textbf{76.14} & 70.46 & 66.29 & 5.21 & \textbf{5.40} & 4.96 & \textbf{67.85} & 66.83 & 63.67 \\
\bottomrule
\end{tabular}
}
\caption{Chunk-TopK results (\%) of various query types on law scenario in Chinese. Seven query types are evaluated: Factual Question (FQ), Information Integration Question (IIQ), Numerical Comparison Question (NCQ), Temporal Sequence Question (TSQ), Multi-hop Reasoning Question (MRQ), Summarization Question (SQ), Unanswerable Question (UQ).}
\label{tab:chunk-topk_performance_law_zh}
\end{table*}

\begin{table*}[t]
\centering
\small
\resizebox{\linewidth}{!}{
\begin{tabular}{lcccccc cccccc}
\toprule
\multirow{2}{*}{Query Type} & 
\multicolumn{6}{c}{\textbf{Retrieval}} & 
\multicolumn{3}{c}{\textbf{Generation}} \\
\cmidrule(lr){2-7} \cmidrule(lr){8-10}
 & \multicolumn{3}{c}{\textbf{Recall ($\uparrow$)}} & 
\multicolumn{3}{c}{\textbf{EIR ($\uparrow$)}} & 
\multicolumn{3}{c}{\textbf{Completeness ($\uparrow$)}} \\
\cmidrule(lr){2-4} \cmidrule(lr){5-7} \cmidrule(lr){8-10}
 & 128-20 & 256-10 & 512-5 & 128-20 & 256-10 & 512-5 & 128-20 & 256-10 & 512-5 \\
\midrule
FQ    & 85.00 & 87.00 & \textbf{90.00} & 1.58 & 2.22 & \textbf{5.15} & 88.00 & 88.00 & \textbf{91.00} \\
IIQ  & \textbf{75.07} & 72.23 & 64.03 & 3.33 & 4.08 & \textbf{5.20} & \textbf{77.50} & 77.00 & 76.17 \\
NCQ   & 52.00 & \textbf{64.50} & 45.17 & 2.34 & 4.17 & \textbf{5.25} & \textbf{54.83} & \textbf{54.83} & 39.33 \\
TSQ   & \textbf{79.33} & 59.17 & 52.00 & \textbf{5.44} & 4.40 & 4.54 & \textbf{58.67} & 45.33 & 40.00 \\
MRQ   & 29.42 & \textbf{30.30} & 19.71 & 3.19 & 7.50 & \textbf{8.93} & \textbf{41.74} & 28.28 & 23.85 \\
SQ    & 18.51 & 25.75 & \textbf{25.90} & 4.23 & 9.03 & \textbf{12.86} & 33.38 & \textbf{36.29} & 34.89 \\
UQ   & \textbf{2.00} & \textbf{2.00} & \textbf{2.00} & \textbf{0.12} & \textbf{0.12} & 0.11 & 79.37 & 83.50 & \textbf{85.83} \\
Avg.  & \textbf{48.76} & 48.71 & 42.69 & 2.89 & 4.50 & \textbf{6.01} & \textbf{62.01} & 59.10 & 55.93 \\
\bottomrule
\end{tabular}
}
\caption{Chunk-TopK results (\%) of various query types on law scenario in English. Seven query types are evaluated: Factual Question (FQ), Information Integration Question (IIQ), Numerical Comparison Question (NCQ), Temporal Sequence Question (TSQ), Multi-hop Reasoning Question (MRQ), Summarization Question (SQ), Unanswerable Question (UQ).}
\label{tab:chunk-topk_performance_law_en}
\end{table*}

\begin{table*}[t]
\centering
\small
\resizebox{\linewidth}{!}{
\begin{tabular}{lcccccc cccccc}
\toprule
\multirow{2}{*}{Query Type} & 
\multicolumn{6}{c}{\textbf{Retrieval}} & 
\multicolumn{3}{c}{\textbf{Generation}} \\
\cmidrule(lr){2-7} \cmidrule(lr){8-10}
 & \multicolumn{3}{c}{\textbf{Recall ($\uparrow$)}} & 
\multicolumn{3}{c}{\textbf{EIR ($\uparrow$)}} & 
\multicolumn{3}{c}{\textbf{Completeness ($\uparrow$)}} \\
\cmidrule(lr){2-4} \cmidrule(lr){5-7} \cmidrule(lr){8-10}
 & 128-20 & 256-10 & 512-5 & 128-20 & 256-10 & 512-5 & 128-20 & 256-10 & 512-5 \\
\midrule
FQ    & 85.00 & \textbf{99.00} & 95.00 & 1.04 & \textbf{1.31} & 0.97 & 80.00 & 94.00 & \textbf{96.00} \\
IIQ  & \textbf{86.83} & 83.83 & 83.50 & 1.95 & \textbf{2.13} & 1.61 & \textbf{87.00} & 83.00 & 79.00 \\
NCQ   & \textbf{83.39} & 70.20 & 70.61 & \textbf{2.90} & 2.87 & 2.39 & \textbf{84.33} & 71.33 & 73.00 \\
TSQ   & \textbf{94.00} & \textbf{94.00} & 87.67 & \textbf{2.72} & 3.21 & 2.35 & 87.33 & \textbf{88.00} & 80.33 \\
MRQ   & 83.36 & 80.90 & \textbf{89.66} & 4.46 & \textbf{4.64} & 4.44 & 67.55 & 62.50 & \textbf{67.73} \\
SQ    & \textbf{85.10} & 77.96 & 83.42 & 6.27 & \textbf{6.32} & 5.98 & \textbf{67.63} & 62.52 & 63.15 \\
UQ   & \textbf{2.00} & 0.00 & \textbf{2.00} & \textbf{0.03} & 0.00 & \textbf{0.03} & \textbf{64.67} & 60.67 & 63.33 \\
Avg.  & \textbf{74.24} & 72.27 & 73.12 & 2.77 & \textbf{2.93} & 2.54 & \textbf{76.93} & 74.57 & 74.65 \\
\bottomrule
\end{tabular}
}
\caption{Chunk-TopK results (\%) of various query types on medical scenario in Chinese. Seven query types are evaluated: Factual Question (FQ), Information Integration Question (IIQ), Numerical Comparison Question (NCQ), Temporal Sequence Question (TSQ), Multi-hop Reasoning Question (MRQ), Summarization Question (SQ), Unanswerable Question (UQ).}
\label{tab:chunk-topk_performance_medical_zh}
\end{table*}

\begin{table*}[t]
\centering
\small
\resizebox{\linewidth}{!}{
\begin{tabular}{lcccccc cccccc}
\toprule
\multirow{2}{*}{Query Type} & 
\multicolumn{6}{c}{\textbf{Retrieval}} & 
\multicolumn{3}{c}{\textbf{Generation}} \\
\cmidrule(lr){2-7} \cmidrule(lr){8-10}
 & \multicolumn{3}{c}{\textbf{Recall ($\uparrow$)}} & 
\multicolumn{3}{c}{\textbf{EIR ($\uparrow$)}} & 
\multicolumn{3}{c}{\textbf{Completeness ($\uparrow$)}} \\
\cmidrule(lr){2-4} \cmidrule(lr){5-7} \cmidrule(lr){8-10}
 & 128-20 & 256-10 & 512-5 & 128-20 & 256-10 & 512-5 & 128-20 & 256-10 & 512-5 \\
\midrule
FQ    & \textbf{98.67} & 91.67 & 80.00 & \textbf{2.61} & 2.24 & 1.70 & \textbf{90.00} & \textbf{90.00} & 88.00 \\
IIQ  & \textbf{93.27} & 80.60 & 81.77 & \textbf{4.93} & 3.27 & 4.28 & 83.00 & 76.00 & \textbf{88.00} \\
NCQ   & \textbf{90.50} & 80.50 & 44.67 & \textbf{2.22} & 1.49 & 0.98 & \textbf{81.00} & 74.33 & 40.67 \\
TSQ   & 81.00 & \textbf{97.00} & 92.00 & \textbf{2.82} & 2.53 & 2.81 & 66.00 & \textbf{73.33} & 72.67 \\
MRQ   & \textbf{66.17} & 64.16 & 40.46 & \textbf{5.83} & 5.23 & 2.64 & 51.22 & \textbf{55.38} & 43.20 \\
SQ    & 57.42 & \textbf{64.35} & 50.29 & 13.74 & \textbf{14.00} & 11.47 & 52.96 & \textbf{59.79} & 52.37 \\
UQ   & 0.00 & 0.00 & 0.00 & 0.00 & 0.00 & 0.00 & 96.00 & 90.00 & \textbf{98.00} \\
Avg.  & \textbf{69.57} & 68.33 & 55.60 & \textbf{4.59} & 4.11 & 3.41 & \textbf{74.31} & 74.12 & 68.99 \\
\bottomrule
\end{tabular}
}
\caption{Chunk-TopK results (\%) of various query types on medical scenario in English. Seven query types are evaluated: Factual Question (FQ), Information Integration Question (IIQ), Numerical Comparison Question (NCQ), Temporal Sequence Question (TSQ), Multi-hop Reasoning Question (MRQ), Summarization Question (SQ), Unanswerable Question (UQ).}
\label{tab:chunk-topk_performance_medical_en}
\end{table*}

\section{Examples}
\label{sec:appendix}

\begin{table*}[h]
\centering
\small
\label{tab:question_type}
\begin{tabular}{p{4cm}p{9.1cm}}
\toprule
\textbf{Question Type} & \textbf{Definition} \\
\midrule
\multicolumn{2}{c}{\textbf{Single-document QA}} \\
\midrule
Factual                & Questions targeting specific details within a reference (e.g., a company's profit in a report, a verdict in a legal case, or symptoms in a medical record) to test RAG's retrieval accuracy.  \\
\midrule
Summarization          & Questions that require comprehensive answers, covering all relevant information, to mainly evaluate the recall rate of RAG retrieval. \\
\midrule
Multi-hop Reasoning    & Questions involve logical relationships among events and details within a document, forming a reasoning chain to assess RAG's logical reasoning ability. \\
\midrule
\multicolumn{2}{c}{\textbf{Multi-document QA}} \\
\midrule
Information Integration & Questions that need information from two documents combined, typically containing distinct information fragments, to test cross-document retrieval accuracy. \\
\midrule
Numerical Comparison    & Questions requiring RAG to find and compare data fragments to draw conclusions, focusing on the model's summarizing ability. \\
\midrule
Temporal Sequence       & Questions requiring RAG to determine the chronological order of events from information fragments, testing the model's temporal reasoning skills. \\
\midrule
\multicolumn{2}{c}{\textbf{Unanswerable Questions}} \\
\midrule
Unanswerable            & Questions arise from potential information loss during the schema-to-article generation, where no corresponding information fragment exists, or the information is insufficient for an answer. \\
\bottomrule
\end{tabular}
\caption{DragonBall Dataset question types and their definitions}
\label{tab:query_type}
\end{table*}

\begin{figure*}[!htbp]
\centering
\begin{tcolorbox}[colback=green!2!white,colframe=gray!50!green]
\begin{minipage}{\linewidth}
\small
% \begin{minted}[fontsize=\footnotesize]{json}
\begin{lstlisting}[language=python]
{
 "courtAndProcuratorate": {
  "court": "",
  "procuratorate": ""
 },
 "chiefJudge": "",
 "judge": "",
 "clerk": "",
 "defendant": {
  "name": "",
  "gender": "",
  "birthdate": "",
  "residence": "",
  "ethnicity": "",
  "occupation": ""
 },
 "defenseLawyer": {
    "name": "",
    "lawFirm": ""
 },
 "caseProcess": {
   "Case Filing and Investigation": {
    "date": ""
   },
   "Detention Measures Taken": {
    "date": ""
   },
   "Criminal Detention": {
    "date": ""
   },
   "Arrest": {
    "date": ""
   }
 },
 "criminalFacts": {
  "Crime Name": {
   "details": [
    {
     "timePeriod": "",
     "behavior": "",
     "evidence": ""
    }
   ]
  }
 },
 "legalProcedure": {
  "judgmentDate": "",
  "judgmentResult": {
   "Crime Name": {
    "sentence": "",
    "sentencingConsiderations": ""
   }
  }
 }
}
\end{lstlisting}

\end{minipage}
\end{tcolorbox}
\caption{A schema example of Law scenario before refinement.}
\label{fig:qar-schema-before-refine-example}
\end{figure*}

\begin{figure*}[!htbp]
\centering
\begin{tcolorbox}[colback=green!2!white,colframe=gray!50!green]
\begin{minipage}{\linewidth}
\small
% \begin{minted}[fontsize=\footnotesize]{json}
\begin{lstlisting}[language=python]
{
 "courtAndProcuratorate": {
  "court": "",
  "procuratorate": ""
 },
 "chiefJudge": "",
 "judge": "",
 "clerk": "",
 "defendant": {
  "name": "",
  "gender": "",
  "birthdate": "",
  "residence": "",
  "ethnicity": "",
  "occupation": ""
 },
 "defenseLawyer": {
    "name": "",
    "lawFirm": ""
 },
 "caseProcess": [
   {
    "event": "Case Filing and Investigation",
    "date": ""
   },
   {
    "event": "Detention Measures Taken",
    "date": ""
   },
   {
    "event": "Criminal Detention",
    "date": ""
   },
   {
    "event": "Arrest",
    "date": ""
   }
 ],
 "criminalFacts": [
  {
   "crimeName": "",
   "details": [
    {
     "timePeriod": "",
     "behavior": "",
     "evidence": ""
    }
   ]
  }
 ],
 "legalProcedure": {
  "judgmentDate": "",
  "judgmentResult": [
   {
    "crimeName": "",
    "sentence": "",
    "sentencingConsiderations": ""
   }
  ]
 }
}
\end{lstlisting}

\end{minipage}
\end{tcolorbox}
\caption{A schema example of Law scenario after refinement.}
\label{fig:qar-schema-example}
\end{figure*}

\begin{figure*}[!htbp]
\centering
\begin{tcolorbox}[colback=green!2!white,colframe=gray!50!green]
\begin{minipage}{\linewidth}
\small
% \begin{minted}
\begin{lstlisting}[language=python]
{
    "courtAndProcuratorate": {
        "court": "Ashton, Clarksville, Court",
        "procuratorate": "Ashton, Clarksville, Procuratorate"
    },
    "chiefJudge": "M. Gray",
    "judge": "H. Torres",
    "clerk": "A. Brown",
    "defendant": {
        "name": "J. Gonzalez",
        "gender": "female",
        "birthdate": "15th, June, 1999",
        "residence": "53, Bayside street, Clarksville",
        "ethnicity": "Hispanic",
        "occupation": "Senior Inspector, Clarksville Tax Department"
    },
    "defenseLawyer": {
        "name": "M. Smith",
        "lawFirm": "Clarksville Legal Associates"
    },
    "caseProcess": [
        {
            "event": "Case Filing and Investigation",
            "date": "1st March 2023"
        },
        {
            "event": "Detention Measures Taken",
            "date": "5th March 2023"
        },
        {
            "event": "Criminal Detention",
            "date": "10th March 2023"
        },
        {
            "event": "Arrest",
            "date": "12th March 2023"
        }
    ],
    "criminalFacts": [
        {
            "crimeName": "Crime of Bending the Law for Personal Gain",
            "details": [
                {
                    "timePeriod": "January 2022 - December 2022",
                    "behavior": "J. Gonzalez utilized her position as Senior Inspector in ...",
                    "evidence": "Email correspondences between J. Gonzalez and ..."
                }
                ...
            ]
        }
    ],
    "legalProcedure": {
        "judgmentDate": "15th May 2023",
        "judgmentResult": [
            {
                "crimeName": "Crime of Bending the Law for Personal Gain",
                "sentence": "5 years of fixed-term imprisonment",
                "sentencingConsiderations": "The defendant's position of trust ..."
            }
        ]
    }
}
\end{lstlisting}

\end{minipage}
\end{tcolorbox}
\caption{A config example of Law scenario.}
\label{fig:qar-config-example}
\end{figure*}

\begin{figure*}[!htbp]
\centering
\begin{tcolorbox}[colback=green!2!white,colframe=gray!50!green, listing options={breaklines=true}]
\begin{minipage}{\linewidth}
\small
% \begin{minted}[fontsize=\footnotesize, breaklines]{json}
\begin{lstlisting}[language=python]

"content":"Ashton, Clarksville Court
Case No. XYZ12345

                                **IN THE COURT OF ASHTON, CLARKSVILLE**
                                    **CRIMINAL JUDGMENT**

**COURT AND PROSECUTOR INFORMATION**

**Court:**
Ashton, Clarksville Court

**Prosecutorate:**
Ashton, Clarksville Procuratorate

**Chief Judge:**
M. Gray

**Judge:**
H. Torres

**Clerk:**
A. Brown

**DEFENDANT AND DEFENSE LAWYER INFORMATION**

**Defendant:**
Name: J. Gonzalez  
Gender: Female  
Birthdate: 15th June 1999  
Residence: 53 Bayside Street, Clarksville  
Ethnicity: Hispanic  
Occupation: Senior Inspector, Clarksville Tax Department

**Defense Lawyer:**
Name: M. Smith  
Law Firm: Clarksville Legal Associates  

**CASE PROCEDURES**

The case against J. Gonzalez commenced with an investigation following a suspicious tip received by the Ashton, Clarksville Procuratorate on 1st March 2023. The investigation revealed substantial evidence implicating the defendant in the Crime of Bending the Law for Personal Gain. Consequently, J. Gonzalez was taken into detention on 5th March 2023. Criminal detention was applied on 10th March 2023, and the defendant was formally arrested on 12th March 2023.

**CASE STATEMENT**

The Crime of Bending the Law for Personal Gain by the defendant, J. Gonzalez, occurred over a span of one year, from January 2022 to December 2022. During this period, J. Gonzalez exploited her position as a Senior Inspector within the Clarksville Tax Department to manipulate tax audits, reports, and reduce penalty fees for several conspiring local businesses in exchange for substantial financial bribes. This court will detail the pertinent events chronologically to provide a comprehensive understanding of the criminal activities committed.

**Charge:**
Crime of Bending the Law for Personal Gain as per Article 397 of the applicable law.

**EVIDENCE DESCRIPTION**

**1. January 2022 - December 2022: Manipulation of Tax Audits in Exchange for Bribes**

During the year of 2022, J. Gonzalez engaged in illicit activities using her privileged position. Emails confirmed numerous correspondences between J. Gonzalez and various local business owners. These emails explicitly outlined her agreement to manipulate tax audits and financial reports for monetary compensation. Bank statements revealed a series of significant transactions amounting to $125,000 deposited into an account owned by J. Gonzalez from suspicious sources. Testimonies from several business owners corroborated these findings, revealing a consistent pattern of bribery and exploitation.

...

**Date of Judgment:**
15th May 2023

**___**
M. Gray, Chief Judge
**___**
H. Torres, Judge
**___**
A. Brown, Clerk"


\end{lstlisting}

\end{minipage}
\end{tcolorbox}
\caption{A document example of Law scenario.}
\label{fig:document-doc-example}
\end{figure*}

\begin{figure*}[!htbp]
\centering
\begin{tcolorbox}[colback=green!2!white,colframe=gray!50!green]
\begin{minipage}{\linewidth}
\small
% \begin{minted}[fontsize=\footnotesize, breaklines]{json}
\begin{lstlisting}[language=python]
{
    "qa_fact_based": [
        {
            "Question Type": "Factual Question",
            "Question": "According to the court judgment of Ashton, Clarksville, Court, what was the judgment date?",
            "ref": [
                "Date of Judgment: 15th May 2023"
            ],
            "Answer": "15th May 2023."
        }
    ],
    "qa_multi_hop": [
        {
            "Question Type": "Multi-hop Reasoning Question",
            "Question": "According to the judgment of Ashton, Clarksville, Court, how many instances of bending the law for personal gain did J. Gonzalez commit?",
            "ref": [
                "The Crime of Bending the Law for Personal Gain by the defendant, J. Gonzalez, occurred over a span of one year, from January 2022 to December 2022.",
                "During this period, J. Gonzalez exploited her position as a Senior Inspector within the Clarksville Tax Department to manipulate tax audits, reports, and reduce penalty fees for several conspiring local businesses in exchange for substantial financial bribes.",
                "In March 2022, J. Gonzalez revised the tax records for Sunrise Construction Inc., drastically reducing their tax liability after receiving a bribe of $50,000.",
                "In exchange for $30,000, J. Gonzalez facilitated the undue reduction of penalty fees levied on Downtown Boutique Ltd. for late tax submissions.",
                "The most egregious of the offenses occurred in November 2022, when J. Gonzalez disclosed sensitive and confidential information about ongoing tax investigations to executives at Riven Pharmaceuticals, securing a bribe of $45,000."
            ],
            "Answer": "According to the judgment, J. Gonzalez committed four instances of bending the law for personal gain: manipulating tax audits and reports, altering tax records, reducing penalty fees, and providing confidential information."
        }
    ],
    "qa_summary": [
        {
            "Question Type": "Summary Question",
            "Summary Content": "Facts of the crime",
            "Question": "According to the judgment of Ashton, Clarksville, Court, summarize the facts of J. Gonzalez's crimes.",
            "ref": [
                "The Crime of Bending the Law for Personal Gain by the defendant, J. Gonzalez, occurred over a span of one year, from January 2022 to December 2022.",
                "During this period, J. Gonzalez exploited her position as a Senior Inspector within the Clarksville Tax Department to manipulate tax audits, reports, and reduce penalty fees for several conspiring local businesses in exchange for substantial financial bribes.",
                "In March 2022, J. Gonzalez revised the tax records for Sunrise Construction Inc., drastically reducing their tax liability after receiving a bribe of $50,000.",
                "In exchange for $30,000, J. Gonzalez facilitated the undue reduction of penalty fees levied on Downtown Boutique Ltd. for late tax submissions.",
                "The most egregious of the offenses occurred in November 2022, when J. Gonzalez disclosed sensitive and confidential information about ongoing tax investigations to executives at Riven Pharmaceuticals, securing a bribe of $45,000."
            ],
            "Answer": "J. Gonzalez, a Senior Inspector at the Clarksville Tax Department, committed the crime of bending the law for personal gain. From January 2022 to December 2022, she manipulated tax audits and reports in exchange for bribes from multiple local businesses. In March 2022, she altered tax records to reduce the tax liability for Sunrise Construction Inc. after receiving $50,000. In August 2022, she reduced penalty fees for late tax submission of Downtown Boutique Ltd. in exchange for $30,000. In November 2022, she provided confidential information about ongoing tax investigations to Riven Pharmaceuticals in exchange for $45,000."
        }
    ]
}
\end{lstlisting}

\end{minipage}
\end{tcolorbox}
\caption{A QRA example of Law scenario.}
\label{fig:qar-QRA-example}
\end{figure*}

\begin{figure*}[!htbp]
\centering
\begin{tcolorbox}[colback=green!2!white,colframe=gray!50!green]
\begin{minipage}{\linewidth}
\small
% \begin{minted}[fontsize=\footnotesize, breaklines]{json}
\begin{lstlisting}[language=python]
{
"prompt":"In this task, you will be given a question and a standard answer. Based on the standard answer, you need to summarize the key points necessary to answer the question. List them as follows:

1. ...
2. ...
   and so on, as needed.

Example:
Question: What are the significant changes in the newly amended Company Law?
Standard Answer: The 2023 amendment to the Company Law introduced several significant changes. Firstly, the amendment strengthens the regulation of corporate governance, specifically detailing the responsibilities of the board of directors and the supervisory board [1]. Secondly, it introduces mandatory disclosure requirements for Environmental, Social, and Governance (ESG) reports [2]. Additionally, the amendment adjusts the corporate capital system, lowering the minimum registered capital requirements [3]. Finally, the amendment introduces special support measures for small and medium-sized enterprises to promote their development [4].
Key Points:

1. The amendment strengthens the regulation of corporate governance, detailing the responsibilities of the board of directors and the supervisory board.
2. It introduces mandatory disclosure requirements for ESG reports.
3. It adjusts the corporate capital system, lowering the minimum registered capital requirements.
4. It introduces special support measures for small and medium-sized enterprises.

Question: Comparing the major asset acquisitions of Huaxia Entertainment Co., Ltd. in 2017 and Top Shopping Mall in 2018, which company's acquisition amount was larger?
Standard Answer: Huaxia Entertainment Co., Ltd.'s asset acquisition amount in 2017 was larger [1], amounting to 120 million yuan [2], whereas Top Shopping Mall's asset acquisition amount in 2018 was 50 million yuan [3].
Key Points:

1. Huaxia Entertainment Co., Ltd.'s asset acquisition amount in 2017 was larger.
2. Huaxia Entertainment Co., Ltd.'s asset acquisition amount was 120 million yuan in 2017.
3. Top Shopping Mall's asset acquisition amount was 50 million yuan in 2018.

Question: Comparing the timing of sustainability and social responsibility initiatives by Meihome Housekeeping Services Co., Ltd. and Cultural Media Co., Ltd., which company initiated these efforts earlier?
Standard Answer: Meihome Housekeeping Services Co., Ltd. initiated its sustainability and social responsibility efforts earlier [1], in December 2018 [2], whereas Cultural Media Co., Ltd. initiated its efforts in December 2019 [3].
Key Points:

1. Meihome Housekeeping Services Co., Ltd. initiated its sustainability and social responsibility efforts earlier.
2. Meihome Housekeeping Services Co., Ltd. initiated its efforts in December 2018.
3. Cultural Media Co., Ltd. initiated its efforts in December 2019.

Question: Based on the 2017 Environmental and Social Responsibility Report of Green Source Environmental Protection Co., Ltd., how did the company improve community relations through participation in charitable activities, community support and development projects, and public service projects?
Standard Answer: Green Source Environmental Protection Co., Ltd. improved community relations through several social responsibility activities. Firstly, in March 2017, the company participated in or funded charitable activities and institutions to support education, health, and poverty alleviation, enhancing the company's social image and brand recognition [1]. Secondly, in June 2017, the company invested in the local community, supporting education, health, and social development projects, deepening its connection with the community and promoting overall community well-being and development [2]. Finally, in August 2017, the company participated in public service projects such as urban greening and public health improvement projects, enhancing the quality of life in the community and promoting sustainable development [3]. These measures enhanced public perception of the company and improved community relations [4].
Key Points:

1. In March 2017, the company participated in or funded charitable activities and institutions to support education, health, and poverty alleviation, enhancing the company's social image and brand recognition.
2. In June 2017, the company invested in the local community, supporting education, health, and social development projects, deepening its connection with the community and promoting overall community well-being and development.
3. In August 2017, the company participated in public service projects such as urban greening and public health improvement projects, enhancing the quality of life in the community and promoting sustainable development.
4. These measures enhanced public perception of the company and improved community relations.

Test Case:
Question: {question}
Standard Answer: {ground_truth} 
Key Points:"

\end{lstlisting}

\end{minipage}
\end{tcolorbox}
\caption{Key points generation prompt.}
\label{fig:key points prompt}
\end{figure*}

\begin{figure*}[!htbp]
\centering
\begin{tcolorbox}[colback=green!2!white,colframe=gray!50!green]
\begin{minipage}{\linewidth}
\small
% \begin{minted}[fontsize=\footnotesize, breaklines]{json}
\begin{lstlisting}[language=python]
{"prompt":"""In this task, you will receive a question, a generated answer, and multiple key points from a standard answer. Please categorize each key point by determining whether it is Relevant, Irrelevant, or Wrong based on the generated answer. For each key point, provide a brief analysis and conclude with one of the following classifications:

[[[Relevant]]] indicates that the generated answer contains key information that is related to and consistent with the key point described in the standard answer.
[[[Irrelevant]]] indicates that the generated answer does not contain or involve information related to the key point in the standard answer.
[[[Wrong]]] indicates that the generated answer contains information related to the key point but it is incorrect or contradicts the standard answer keypoints.

Ensure that each key point is categorized into only one of the three categories. Provide your analysis and conclusion for each key point sequentially.

Example 1:

Question: What is the current situation and future outlook for wealth management fee income?

Generated Answer: In the first three quarters, the bank achieved wealth management fee income of 5.105 billion yuan, down 20.7% year-on-year. This decline was mainly due to the active reduction of non-standard product scales and the decline in fund sales caused by fluctuations in the securities market. To address this situation, the bank plans to continue upgrading its products, teams, professionalism, and technological capabilities. Specific measures include improving asset selection and product screening capabilities, enriching and optimizing product shelves, continuously strengthening the construction of professional teams, and focusing on enhancing team professional skills and middle-office empowerment levels to enhance the endogenous development capabilities of the teams. The future outlook includes achieving more stable and higher-quality development through these measures to cope with the challenges of declining wealth management fee income.

Standard Answer Key Points:
Here are 2 key points
1. Current situation of wealth management fee income: Income in the first three quarters decreased by 21.6% year-on-year.
2. Future measures: Plans to upgrade products and teams to achieve stable development.

Key Point Evaluation:

Key Point 1:
Comparable fragment of the generated answer: "wealth management fee income of 5.105 billion yuan, down 20.7% year-on-year."
Analysis: The generated answer mentions a decrease of 20.7% year-on-year, whereas the standard key point specifies a decrease of 21.6%. The percentage difference makes this information incorrect.
Conclusion: [[[Wrong]]]
Key Point 2:
Comparable fragment of the generated answer: "plans to continue upgrading its products, teams, professionalism, and technological capabilities."
Analysis: The generated answer aligns with the standard key point by detailing plans to upgrade products and teams to achieve more stable development.
Conclusion: [[[Relevant]]]
.... omit three example here

Before you begin the evaluation, please pay attention to the following points:

1. [[[Wrong]]] should only be assigned when there is a specific factual or logical conflict between the key point and the generated answer. If important content is missing, it should be categorized as [[[Irrelevant]]], not [[[Wrong]]]. More special cases should refer to point 5 below.
2. [[[Relevant]]] does not require the generated answer to include all the details. It only needs to contain the key information necessary to answer the question. Not all details are required. We ensure that each key point in the standard answer is typically necessary, although some details might not be important for answering the question. When making judgments, focus only on whether the most important information is included and consistent. Also, identical content in different forms can be considered relevant as long as the core key information is present.
3. Please ensure that the number of key points evaluated matches the number of key points in the standard answer. Each key point must be evaluated; do not skip or over-evaluate any key point.
4. After evaluating the key points, do not repeat your conclusions. Ensure that the total number of classifications - [[[Relevant]]], [[[Wrong]]], and [[[Irrelevant]]] - matches the number of key points in the standard answer.
.... omit three more instruction

Test cases:
Question: {question}
Generated Answer: {prediction}
Standard Answer Key Points:
Here are {key_points_num} key points
{key_points}
Key Point Evaluation:"""
}
\end{lstlisting}

\end{minipage}
\end{tcolorbox}
\caption{Key points evaluation prompt.}
\label{fig:key points prompt}
\end{figure*}

\end{document}